\documentclass[a4paper, 10 pt, conference]{ieeeconf}

\IEEEoverridecommandlockouts                            

\usepackage[utf8]{inputenc} %
\usepackage[T1]{fontenc}    %
\usepackage{hyperref}       %
\usepackage{url}            %
\usepackage{booktabs}       %
\usepackage{amsfonts}       %
\usepackage{nicefrac}       %
\usepackage{microtype}      %
\usepackage{lipsum}
\usepackage[T1]{fontenc}

\usepackage{cite}

\usepackage{amsmath,amssymb}
\usepackage{algorithmic}
\usepackage{graphicx}
\usepackage[table,xcdraw]{xcolor}
\usepackage{subfigure}
\usepackage{textcomp}
\usepackage{gensymb}
\usepackage{epsfig} %
\usepackage{xcolor}

\usepackage{booktabs}
\usepackage{array}
\usepackage{multirow}

\usepackage{ulem}
\usepackage{units}
\newcommand{\subparagraph}{}

\title{\LARGE \bf A Comprehensive Dataset of Grains for Granular Jamming in Soft Robotics: Grip Strength and Shock Absorption}

\author{David Howard$^{1}$, Jack O'Connor$^{1,2}$, Jordan Letchford$^{1,3}$ Therese Joseph$^{1,3}$,\\ Sophia Lin$^{1,4}$, Sarah Baldwin$^{1}$, and Gary Delaney$^{1}$
\thanks{$^{1}$ CSIRO, Australia; contact {david.howard@csiro.au}}
\thanks{$^{2}$ University of Queensland, Australia}
\thanks{$^{3}$ Queensland University of Technology, Australia}
\thanks{$^{4}$ University of Melbourne, Australia}
}

\begin{document}
\maketitle
\thispagestyle{empty}
\pagestyle{empty}

\begin{abstract}
We test grip strength and shock absorption properties of various granular material in granular jamming robotic components. The granular material comprises a range of natural, manufactured, and 3D printed material encompassing a wide range of shapes, sizes, and Shore hardness. Two main experiments are considered, both representing compelling use cases for granular jamming in soft robotics. The first experiment measures grip strength (retention force measured in Newtons) when we fill a latex balloon with the chosen grain type and use it as a  granular jamming gripper to pick up a range of test objects.  The second experiment measures shock absorption properties recorded by an Inertial Measurement Unit which is suspended in an envelope of granular material and dropped from a set height. Our results highlight a range of shape, size and softness effects, including that grain deformability is a key determinant of grip strength, and interestingly, that larger grain sizes in 3D printed grains create better shock absorbing materials.  The data set is publicly available at \url{https://doi.org/10.25919/tgck-2r85}.

\end{abstract}

\begin{keywords}
Soft robotics, Soft gripping, Shock absorbance, Granular jamming
\end{keywords}

 \section{INTRODUCTION}
\label{sec:introduction}

\begin{figure}[t!]
\centering
 \includegraphics[width=0.9\columnwidth]{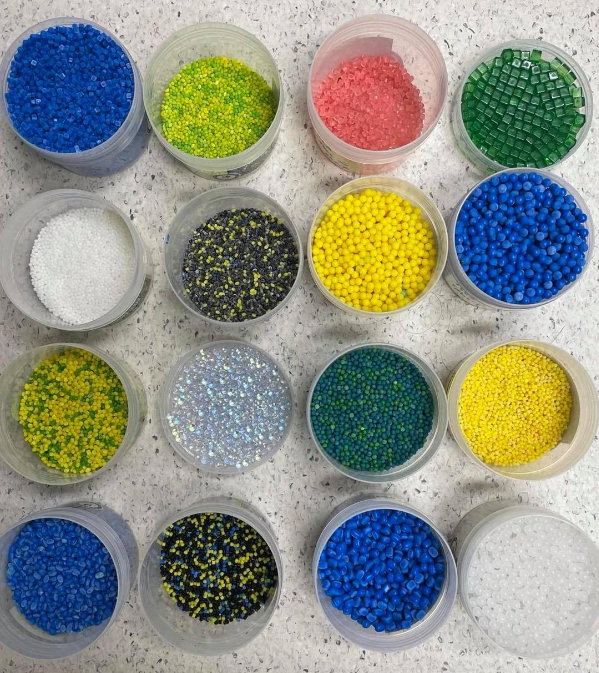}
\caption{Exemplar granular material used in the experiments, encompassing a range of shapes, sizes, materials, and hardness/softness.}
\end{figure}

Variable stiffness \cite{manti2016stiffening} is a key feature of many soft robotic systems.  Selective softening allows for compliance like a pneumatic actuator \cite{xavier2022soft}, whereas selective stiffening permits meaningful force transmission. Amongst the candidate variable stiffness soft robotics technologies, granular jamming occupies a sweet spot in terms of rapidly fast actuation and a large delta between maximum softness and maximum rigidity \cite{fitzgerald2020review}.  Jamming is therefore a popular choice for achieving variable stiffness in soft robotics.

Granular jamming refers to the property of a collection of grains to transition from a fluid-like behaviour to a solid-like behaviour when the constituent grains are forced together, e.g. under vacuum pressure when confined within a membrane.  By far the most prevalent use of this mechanism in soft robotics applications is seen in gripping, where 'universal grippers' provide simple object-agnostic gripping by imprinting the gripper against the target object in the fluid-like state, and subsequently jamming the grains to exert a gripping force on the object \cite{brown2010universal}.  Jamming has a number of additional benefits in soft robotics depending on the specific application, including shock absorption, compliance, and deformation, which has seen them deployed in diverse applications including prosthetics ~\cite{cheng2016prosthetic} and undersea object manipulation ~\cite{licht2017stronger}.

Regardless of their final embodiment, granular soft robotic systems are attractive because of the large number of design variables that can be harnessed to elicit a desired behaviour.  The literature explores the effects of these design variables, as well as supporting methods to more easily access the design space.  The effects of membrane material on gripper performance has been investigated \cite{jiang2014robotic}, as has optimising membrane morphology with a fixed (3D printed) material \cite{howard2022getting}.  Both studies highlight significant performance differences within the design variable ranges covered, and recent work has focused on 3D printing techniques to allow membrane optimisation to become more widespread and freeform \cite{howard2022one}.  

As well as membranes, grains offer the chance to elicit application-specific performance.  This is particularly intriguing, as the specific shape, size and material of an individual grain can have large, potentially unpredictable effects on the bulk properties of the granular structure.   Grains are the most frequently investigated design variable, although most studies consider only a small, sometimes arbitrary set of candidate grains, either manufactured (e.g. plastic or glass spheres, rubber cubes) or natural (e.g. coffee, rice) in origin \cite{jiang2013granular,hauser2018stiffness,chopra2020granular}.  Research has also focused on the use of 3D printing to explore grain shape and size \cite{howard2021shape}, showing not only that grain shape and size are key determinants in granular gripping performance, but also that there is no 'universal' best shape/size combination across the test objects considered.  In other words, grains play a meaningful part in the design of optimal granular grippers.  We also note the use of machine learning to create bespoke grains, either in granular structures generally \cite{miskin2013adapting}, or more recently specifically for soft robotics  \cite{fitzgerald2021evolving}, and later extended to consider mixtures of grains to further expand the potential design space \cite{fitzgerald2022evolving}.

Recent work has highlighted the benefits of grain softness for gripping \cite{gotz2022soft,putzu2019soft}, based on the extra force generated from compressing a soft granular packing under vacuum.  Similar to the majority of the literature, this work only considers a limited range of possible grains, and our paper seeks to expand the range of soft grains that are evaluated.  

Overall, even though most research on granular jamming for soft robotics includes some limited form of grain comparison, to date no research has offered a comprehensive comparison of a range of grain shapes, sizes, and materials.  A 2020 literature review \cite{fitzgerald2020review} shows an average of $<3$ grain comparisons per paper, but a total of 24 grains used across the literature.  A primary motivation for this work is therefore to  offer a holistic view of the impact of grain choice on granular gripping in soft robotics considering a wide range of popular grains. 

Gripping is not the only use case for granular jamming in soft robotics.  Granular systems also exhibit excellent shock absorbance properties, which has a variety of applications in soft robotics including natural terrain locomotion \cite{hauser2016friction}, wearable devices \cite{thompson2015soft}, and damage resistance \cite{steltz2010jamming}.  We therefore include a second experiment that assesses the shock absorbance of the same set of grains.  The majority of the literature is found in the field of granular physics rather than soft robotics.  Impact resistance \cite{ji2013granular} and vibration damping properties \cite{An_2020} have been investigated, as well as the effect of grain shape on stress response \cite{athanassiadis2014particle} and energy dissipation \cite{dragomir2012}, however the area is relatively understudied (especially compared to the amount of work on gripping) and experimental setups are not representative of granular systems that could be used in soft robotics. 

\subsection{Motivation \& Contributions}
An important and open research question in granular jamming for soft robotics is therefore {\it'What is the best grain choice for my granular jamming soft robotic device?'}. To answer this question, we assemble the largest collection of grains to date, and test them on two fundamental application cases: gripping and shock absorbance.  The experimental setups used in each case are simple to recreate, allowing researchers to validate and extend the data set.  The key contributions of this paper are:

\begin{itemize}
\item The largest data set on grains for gripping.
\item The only data set on grains for shock absorbance in a soft robotics context.
\item Analysis of key trends in both cases, providing some guidance for researchers wishing to deploy granular systems into their soft robotics projects.
\end{itemize}

A brief summary of our results shows that shape, size, and material all impact grip strength, and that grain softness is the key differentiator, with soft grains outperforming hard grains.  For shock absorbance, grain size rather than grain softness was the key determinant of performance, and soft grains were outperformed by hard grains.

\section{Candidate grains}

Our data set is comprised of 35 grains, of which 13 are purchased and the rest 3D printed (Table.\ref{tab:grain_types}). Together they cover a range of shapes, sizes, and materials.  Five of the grains are soft, 30 are rigid.  The purchased grains are frequently seen in the literature and therefore those most relevant to the research community.  They were sourced from a range of commercial suppliers.  Printing allows us to span a range of variables (shape, size, softness) at regular intervals and draw out patterns in the data.  

\begin{table}[ht]
\caption{Grain Type Legend}
\label{tab:grain_types}
\resizebox{\columnwidth}{!}{%
\begin{tabular}{ccccc}
\hline
\textbf{Code} & \textbf{Shape} & \textbf{Size} & \textbf{Material} & \textbf{Source} \\ \hline
C 3SVE V & Cube & 3mm SVE & Vero & Inhouse \\ 
E 3SVE V & Ellipsoid & 3mm SVE & Vero & Inhouse \\ 
S 3SVE V & Sphere & 3mm SVE & Vero & Inhouse \\ 
SE 3SVE V & Super Ellipsoid & 3mm SVE & Vero & Inhouse \\ 
C 4SVE V & Cube & 4mm SVE & Vero & Inhouse \\ 
E 4SVE V & Ellipsoid & 4mm SVE & Vero & Inhouse \\ 
S 4SVE V & Sphere & 4mm SBE & Vero & Inhouse \\ 
SE 4SVE V & Super Ellipsoid & 4mm SVE & Vero & Inhouse \\ 
C 5SVE V & Cube & 5mm SVE & Vero & Inhouse \\ 
E 5SVE V & Ellipsoid & 5mm SVE & Vero & Inhouse \\ 
S 5SVE V & Sphere & 5mm SVE & Vero & Inhouse \\ 
SE 5SVE V & Super Ellipsoid & 5mm SVE & Vero & Inhouse \\ 
C 6SVE V & Cube & 6mm SVE & Vero & Inhouse \\ 
E 6SVE V & Ellipsoid & 6mm SVE & Vero & Inhouse \\ 
S 6SVE V & Sphere & 6mm SVE & Vero & Inhouse \\ 
SE 6SVE V & Super Ellipsoid & 6mm SVE & Vero & Inhouse \\ 
C 7SVE V & Cube & 7mm SVE & Vero & Inhouse \\ 
E 7SVE V & Ellipsoid & 7mm SVE & Vero & Inhouse \\ 
S 7SVE V & Sphere & 7mm SVE & Vero & Inhouse \\ 
SE 7SVE V & Super Ellipsoid & 7mm SVE & Vero & Inhouse \\ 
S 3 G & Sphere & 3mm & Glass & Bought \\ 
S 6 G & Sphere & 6mm & Glass & Bought \\ 
S 8 G & Sphere & 8mm & Glass & Bought \\ 
S 3 SP & Sphere & 3mm & Solid Plastic & Bought \\ 
C 3 HP & Cylinder & 3mm & Hollow Plastic & Bought \\
C 7 HP & Cube & 7mm & Hollow Plastic & Bought \\ 
S 7 HP & Sphere & 7mm & Hollow Plastic & Bought \\ 
Rice & Long Grain & - & White Rice & Bought \\
Coffee & Grinds & - & Coffee & Bought \\ %
Polystyrene & Sphere & - & Polystyrene & Bought \\ 
Rubber & Crumb & - & Rubber & Bought \\ 
C 4 AG70 & Cube & 4mm & Agilus70 & Inhouse \\ 
C 4 AG30 & Cube & 4mm & Agilus30 & Inhouse \\ 
SEBS0 & Ellipsoid & - & SEBS0 & Bought \\ 
SEBS30 & Ellipsoid & - & SEBS30 & Bought \\ 
\end{tabular}%
}
\end{table}
\begin{figure}[h!]
\centering
\label{fig:STLgrains}
\includegraphics[width=0.9\columnwidth]{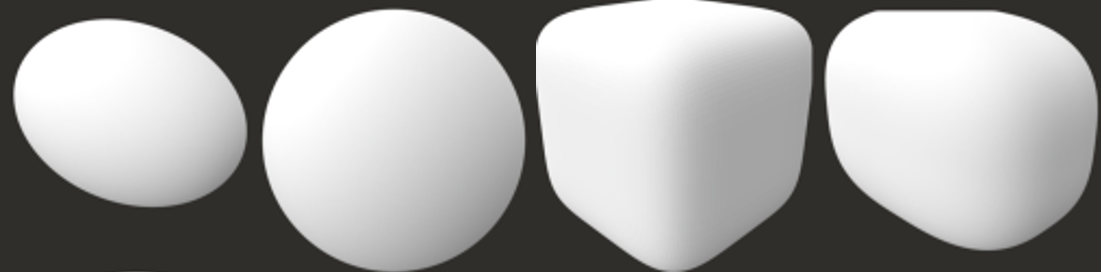}
\caption{CAD files of 3D printed grain shapes: ellipsoid, sphere, cube (superball), superellipsoid.}
\end{figure}

Each grain is given a shorthand code for easy reference, and its shape, size, and material are recorded.  The code for each grain starts with its shape (if any); (C)ube, (E)llipsoid, (S)phere or (SE)SuperEllipsoid as shown in Fig~\ref{fig:STLgrains}.  Irregular shaped grains such as coffee or rubber crumb are not given a shape code, although their approximate shape is recorded.  

For 3D printed grains, the shape code relates to a precise shape.  Formulation of the shapes and their selection to generate diverse behaviour is given in \cite{howard2021shape} -- see also \cite{delaney2010packing}.  In brief, 3D printed grains are chosen from their ability to create diverse behaviours as predicted by DEM modelling. Printed grains are parameterised superquadrics following (1).
\begin{equation}
  (x/a)^{m} + (y/b)^{m} + (z/c)^{m} = 1.
\end{equation}
The specific shapes are {\it sphere} ($m$=2, $a$=$b$=$c$=1), {\it ellipsoid} ($m$=2, $a$=1, $b$=$c$=0.65), {\it cube} ($m$=5, $a$=$b$=$c$=1) and {\it superellipsoid} ($m=3$, $a$=1 $b$=0.75 $c$=0.6) (Fig.~\ref{fig:STLgrains}).  Each shape was made in 3mm, 4mm, 5mm, 6mm, and 7mm SVE.

The second code shows the size of the grain in $mm$.  Irregular size grains do not have their shape coded.  For printed grains, the suffix {\it SVE} (Sphere Volume Equivalent) means the grain is printed to match the volume of an equivalently sized sphere: A 4SVE cube's volume equals that of 4mm diameter sphere.

Finally, material type is coded.  Hard printed grains are fabricated in (V)ero material, and soft printed grains using either Agilus or an Agilus/Vero mix with corresponding Shore-A value (e.g., AG30), on a Stratasys Connex3 Objet 500 polyjet printer following the CAD files in Fig~\ref{fig:STLgrains}. All grains are printed with a layer height of 16 microns. Purchased grains include glass (G), solid and hollow HDPE plastic (SP/HP), rubber, polystyrene, SEBS, rice, and coffee.

\section{Experiment 1: Gripping}

\subsection{Setup}

A 12.5cm latex balloon is filled with each grain type in turn, and secured to a Dremel drill press stand via a 3D printed adapter. A Rocker 300 vacuum pump is then connected to the balloon gripper through the adapter via a filter and silicon tube.  The test object is screwed into a Zemic H3-C3-25kg-3B load cell via a 3D printed thread.  The load cell attaches to the base of the drill press stand via a metal mounting plate.  A flat 3D printed platform (50mm x 50mm) is attached between the test object and load cell to replicate the action of picking an object off a flat surface\footnote{Granular grippers typically require the object to be pushed onto a surface prior to executing a grip}. The test objects are selected to provide a diverse range of challenges, including objects such as the coin that granular grippers struggle to grip \cite{howard2021shape}.

\begin{figure}[ht]
\centering
\subfigure{\includegraphics[width=0.9\columnwidth]{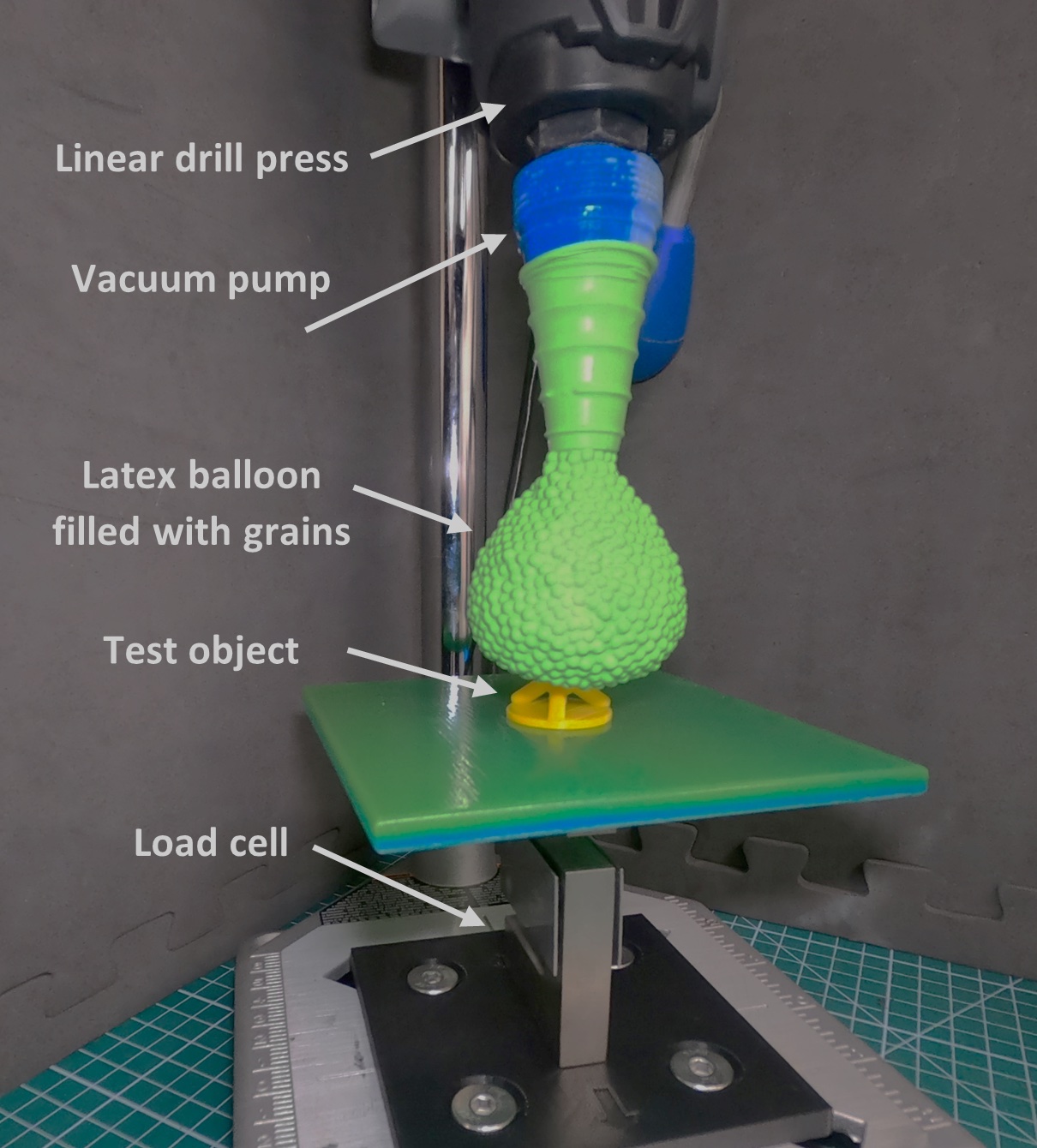}}\\
\subfigure{\includegraphics[width=0.9\columnwidth]{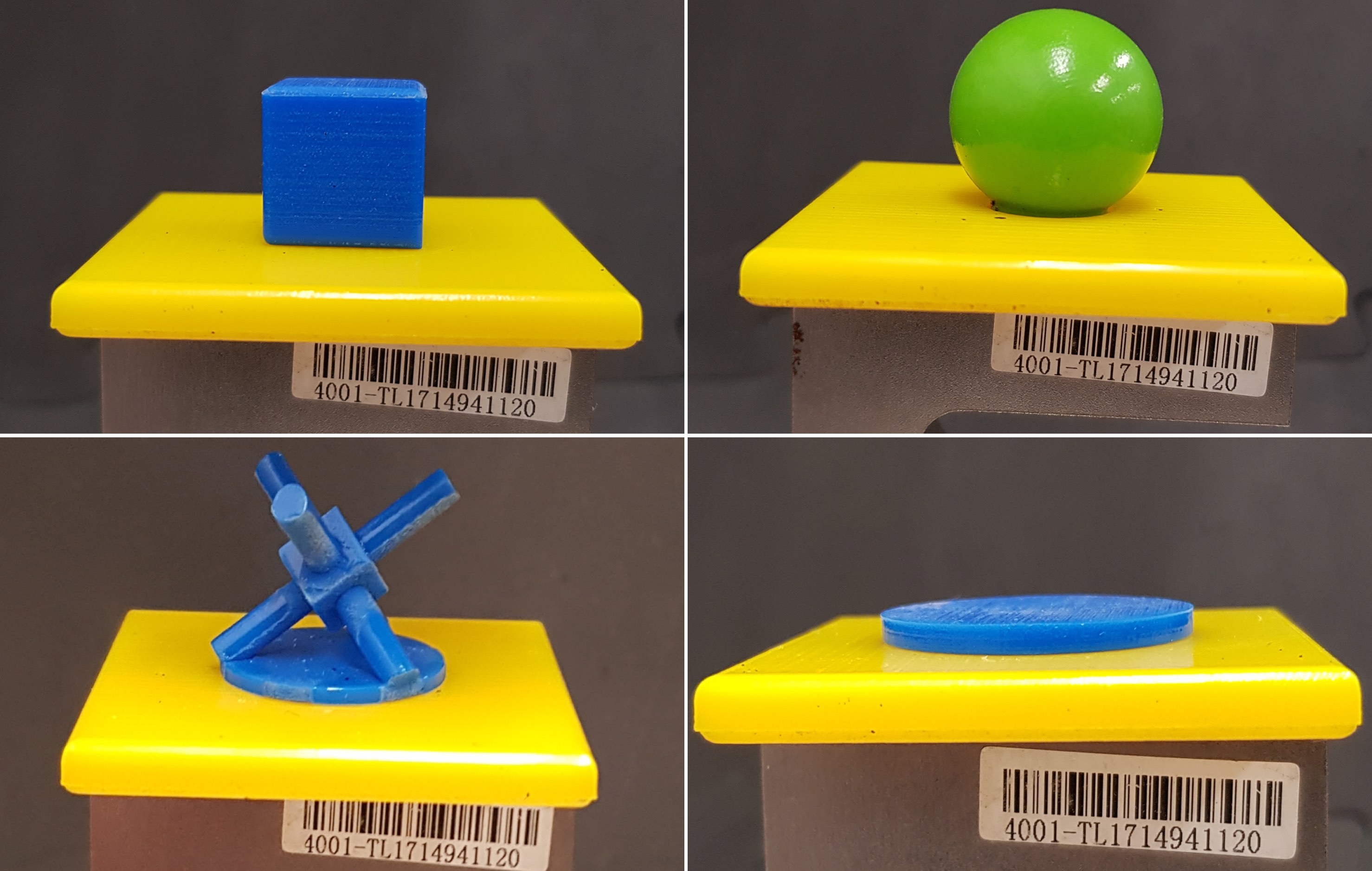}}
\caption{Experimental setup for gripping.  Top: A latex balloon filled with a selected grain type is lowered via a linear drill press onto a test object.  The object is 3D printed with a thread that screws into a load cell.  Bottom: The four test objects (clockwise from top-left) Cube, Ball, Coin, Star.  Each object is 20mm$^3$.}
\end{figure}

To execute a test, the gripper is lowered onto the object in an unjammed state until the lever was fully depressed.  The vacuum pump is activated, transitioning the gripper into a jammed state and gripping the object. The gripper is then slowly raised until it completely clears the test object. A regulator maintains pressure at -60kPa.  The vacuum was then deactivated and the gripper manually reset by 10s of shaking \footnote{Heuristically selected based on initial testing to fully reset the gripper internally}. To ensure representative results, each test is repeated 10 times per test object.

\subsection{Gripping Results}

\begin{figure*}[t!]
\label{fig:gripresults}
\centering
\subfigure[]{\includegraphics[width=\columnwidth]{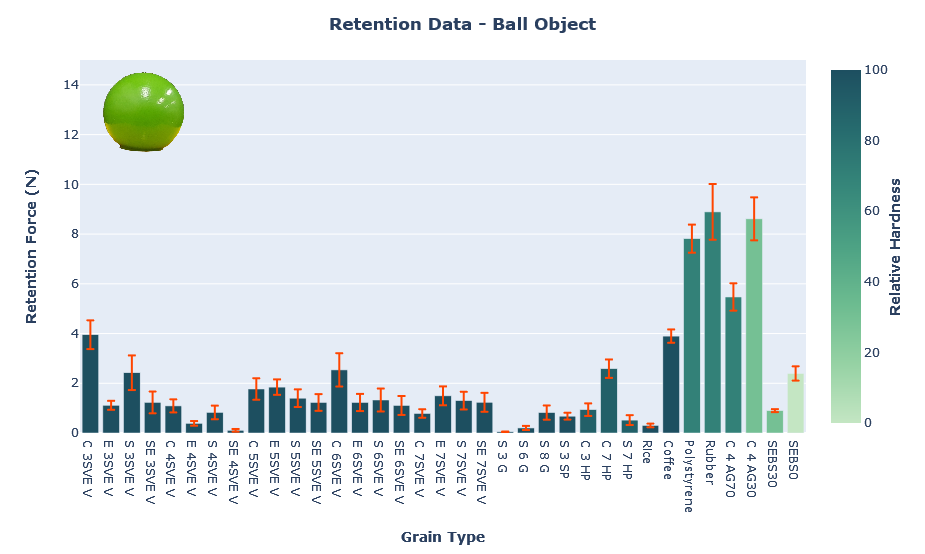}}%
\subfigure[]{\includegraphics[width=\columnwidth]{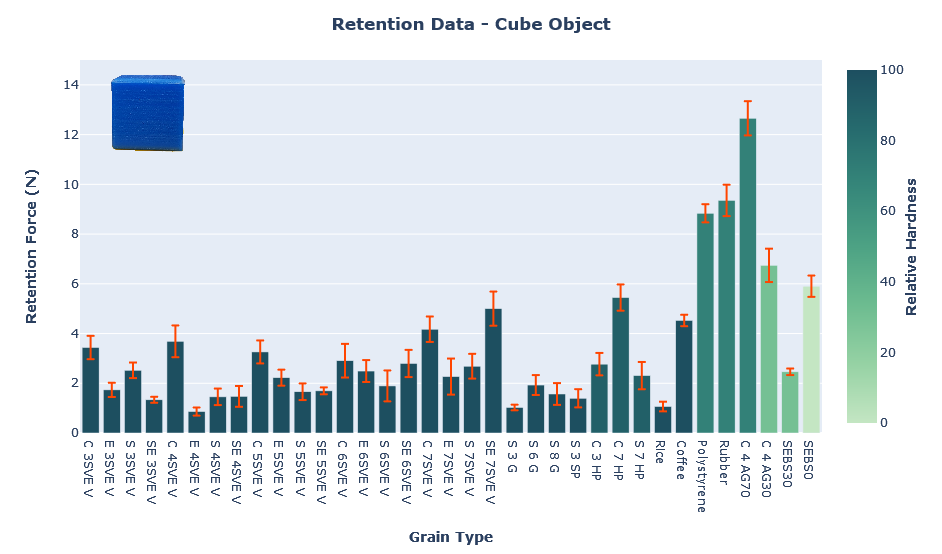}}\\
\subfigure[]{\includegraphics[width=\columnwidth]{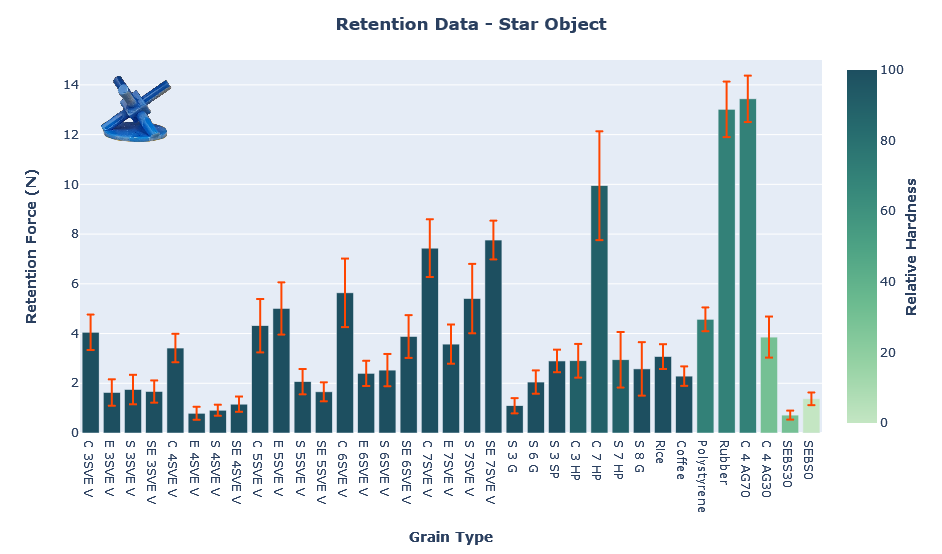}}%
\subfigure[]{\includegraphics[width=\columnwidth]{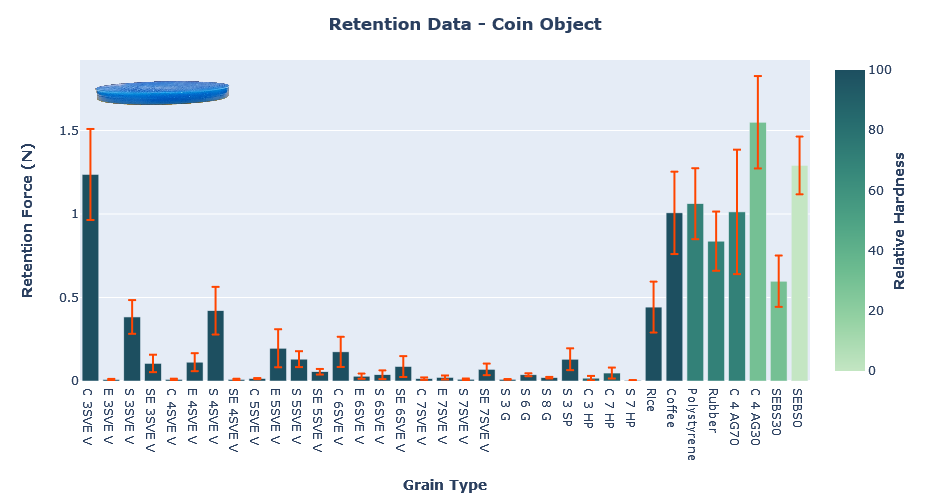}}%
\caption{Retention Force vs. Grain Type for gripping tests, ordered by test object: (a) Ball (b) Cube (c) Star (d) Coin. Bars are coloured according to grain hardness; dark green are hard, lightest green is Shore 0.}
\end{figure*}

Gripping results are seen in Fig.~\ref{fig:gripresults}, split by test object.  Results are summarised in Table~\ref{tab:results_summary}.  Statistical significance is assessed via a Mann-Whitney U-test with a p-value $<0.05$ and is provided in Appendix Tables~\ref{tab:cube_grip},~\ref{tab:star_grip},~\ref{tab:ball_grip}, and ~\ref{tab:coin_grip}.

\subsubsection{Effects of Test Objects}
Each grain shape generally displayed increased performance on the star test object and decreased performance on the coin. The star offers numerous overhangs due to its geometry, which the gripper can seal around to create a strong grip, resulting in a higher gripping force.  The complex geometry of the star object supported multiple different possible contact points during a grip, causing varying grip strengths between the configurations.  

In terms of the percentage variability of the gripping force from one grip to another on the same object, the cube has the most consistent gripping force, followed by star and ball, and then the coin shows the greatest variability.  Despite relatively small gripping forces, the coin grips occasionally outright failed, contributing significantly to the measured variability in the grip strength.

\subsection{Size Effects}
On average, our small solid grains (<4mm) had the lowest grip performance (mean <4N). Of these, the ball and coin target object produced the lowest gripping performance with an average retention force of less than 2N and 0.5N respectively.  There is a trend for larger grain sizes to produce greater retention forces on these objects, particularly the cube and super ellipsoid grain shapes of 5-7mm. It is likely that these combinations of grain size and shape produce complimentary jamming geometries with the cube and star object in particular.  There are strong size dependencies observed, e.g., 3SVE cubes perform well on the coin object, however 4SVE cubes perform poorly. The 3SVE grains are small enough to easily achieve a set of strong face-on-edge geometrically jammed contacts with around coin, which is not the case for the larger 4SVE cubes.

\subsection{Shape Effects}
It is interesting to note that unconventional grain shapes that are highly underrepresented in the literature, despite demonstrating improved performance in many applications, e.g. super-ellipsoids exhibit particularly high gripping forces on the on cube and star objects.  This indicates that printing of custom grain geometries is a promising route towards bespoke jamming systems with tailored performance \cite{howard2021shape}.

The cube grain shape consistently performs well across test objects, sizes and relative hardness, although this increased performance also sees an increase in the variability of the measured gripping strength. This high performance is attributed to large number of face-on-face contacts between grains in the bulk, increasing the internal strength of the granular material and its resistance to deformation during the pull off process.  Cubes can also geometrically jam against certain test objects, e.g., within in the nooks of the star and flush to the edges of the cube and coin, providing very strong grips.  However, this does not always happen, resulting in higher variability in gripping strength compared to simple spherical grain shapes.  

For solid grains, the combination of grain shape and size, and object shape and size both had significant impacts on gripping performance, with no discernible rule for all cases. Thus we find that the selection of the highest performing grain is highly application dependent, with significant potential for optimisation of performance for a particular target object's shape and size.

\subsection{Material Effects}

We have considered grains with a range of different stiffnesses and find that there is a increase in performance with soft grains across all target objects. 
The soft grains performed well across all tests, and also gave the overall highest performance for each of the target object shapes. This is likely due to their ability to easily conform to each other and the test object -- the theory behind this effect is covered in \cite{gotz2022soft}. Furthermore,  rubber-like properties  provide a higher friction coefficient, enabling the granular material to form a stronger bond with neighbouring grains and form a tight-packing matrix. Further research is required around the hystersis and shape recovery ability of softer cubes, especially for grains like polystyrene which might undergo structural changes when compressed repeatedly.

The overall best performing grain shape for each target object was the Agilus70 4mm cube for the star (13.4N) and cube (12.6N) targets, the rubber crumb for the  Ball (8.8N), and the Agilus30 4mm cube for the coin (1.5N). Comparatively, the best solid grain performance for each test object is the 7mm SVE super ellipsoid for the star (7.7N) and cube(5.0N), then the 3mm SVE cube for the ball (3.9N) and coin (1.2N). 

Of the hard plastics grains, the hollow plastic cube grains consistently out-perform the other hollow plastic grains, and most of the solid plastics also peaking with an average of 9.95N for 7mm hollow plastic cube grains on the star object, compared to an average max of 5N between sphere plastic grains. This continues to suggest that some material compliance within the grains may be advantageous for enhancing the granular jamming action and resulting retention performance. This point is highlighted when directly comparing the range of 4mm SVE cube grains, with Agilus70 and Agilus30 consistently outperforming Vero, with maximum retention seen in Agilus70 cubes on the star object (13.44N) compared to lowest seen in Vero (3.42N).

All flexible grains performed well, albeit with comparatively large performance variability compared to solid grains, except for the softest grain which has the performance variability of the soft grains. Soft grains supplied the best performance for each object with polystyrene for the ball (7.82N), Agilus70 (12.66N) for the cube and star (13.44N) and Agilus 30 for the coin (1.55N). Because soft grains are so compliant, they may not pack and jam in a consistent manner. These findings suggests that soft grains can perform favourably to manipulate different object shapes. Despite their larger variability suggesting less repeatable performance results, their unique material properties compliment the nature of jamming actuation to result in reliably larger average retention force and grip performance.

\section{Experiment 2: Shock Absorption}

\subsection{Setup}

\begin{figure}[ht]
\centering
\label{fig:shocklabel}
\includegraphics[width=0.75\columnwidth]{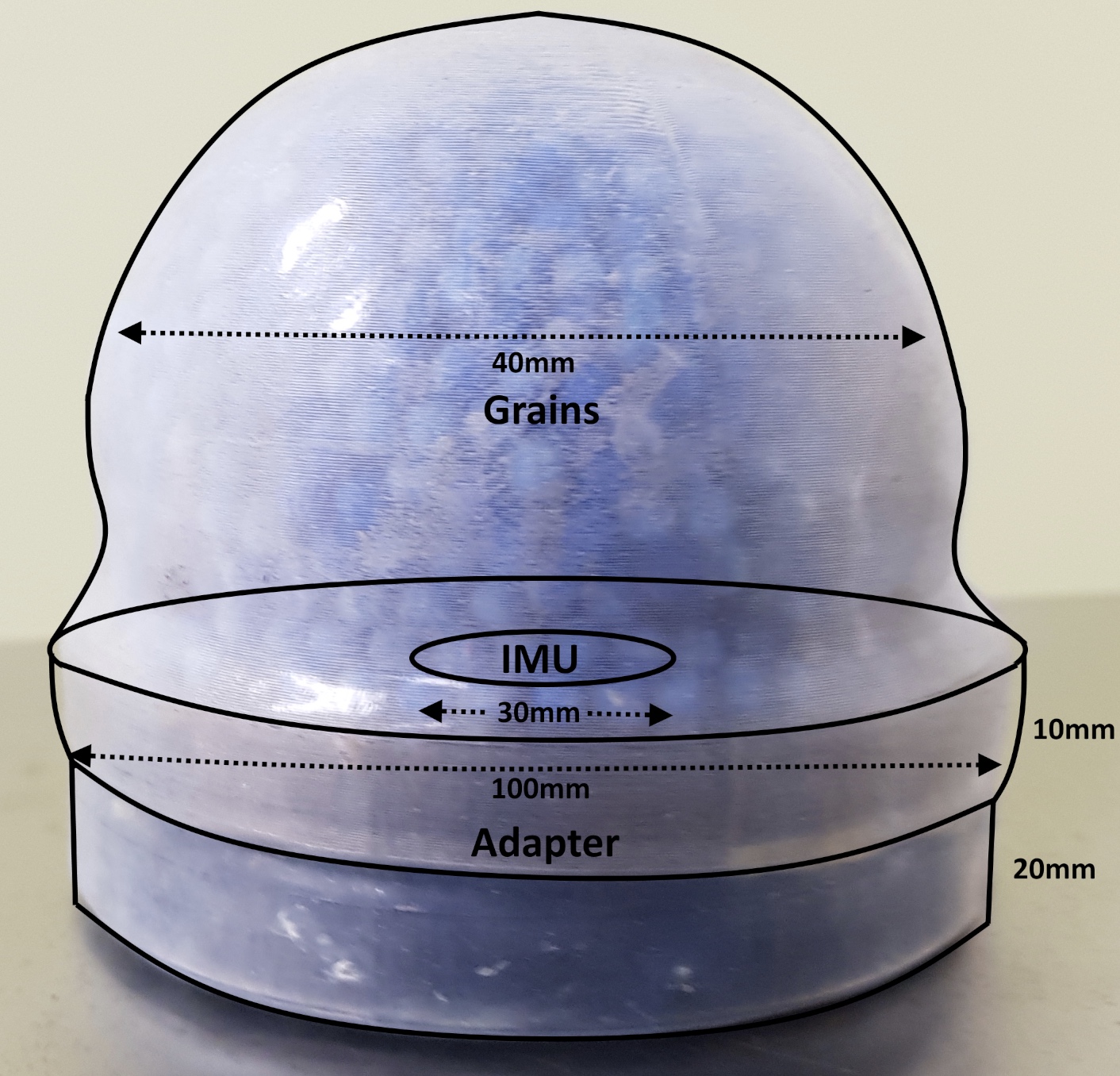}
\caption{Experimental set up for shock absorption tests.  A spherical silicone membrane is filled with grains to a constant volume. A bluetooth IMU is inserted into in a 3D printed adapter and the membrane is tightly compression fitted over it.  The device faces grains-down and is dropped from a height of 30cm, measured from a custom platform, and shock experienced by the IMU is recorded.}
\end{figure}

The shock test setup was designed to replicate the type of fall that a robot may be subjected to, e.g., when exploring an unknown environment or performing a search and rescue mission. A Radioland Technology NRF52832 Beacon Bluetooth Inertial Measurement Unit (IMU) was inserted into a custom 3D printed adapter of 100mm diameter.  A cast silicon sleeve was filled with the chosen grain type to a constant volume and stretched over the adapter (Fig.\ref{fig:shocklabel}). The sleeve was shaped as a pinched 40mm diameter demisphere to provide a tight compression fit over the adapter whilst being re-usable, and the IMU was placed centred in the horizontal plane 40mm from the top of the membrane. 

Once the wireless IMU had been initialised, the sleeve was manually lifted to a marked height of 30cm, orientated to point downwards through placement on a custom test platform, and dropped. Each test was was repeated 10 times per grain type, and because unjammed structures are known to provide maximum shock absorption, only the unjammed state of each grain was tested.  

Each reading is calculated by taking the largest delta in G between the start and end of the experiment, which is typically observed as a spike at impact time.  Low values are preferred, indicating that the IMU experienced lower forces due to dissipation effects from the granular material.

\subsection{Shock Absorbance Results}

\begin{figure}[ht]
\label{fig:shockresults}
\centering
\subfigure[]{
\includegraphics[width=\columnwidth]{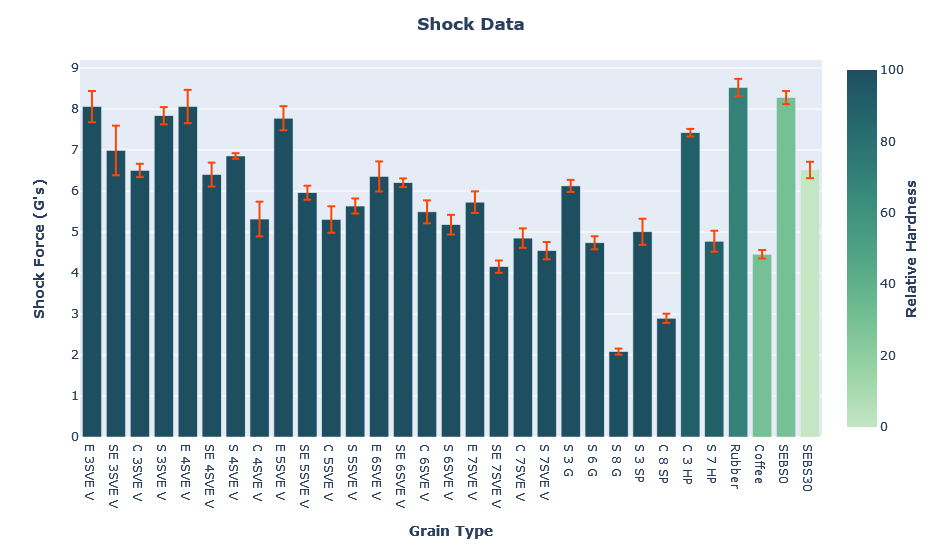}}
\subfigure[]{
\includegraphics[width=\columnwidth]{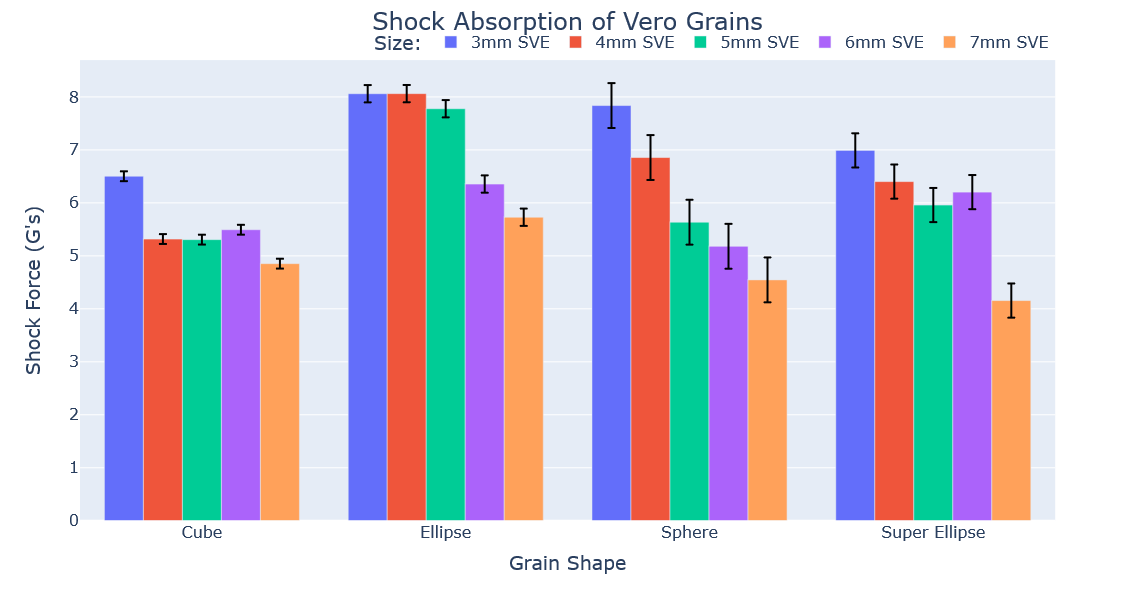}}
\caption{(a) Shock Absorption properties of each grain type in units of G (lower is better). Bars are coloured according to grain hardness, with darker greens representing harder materials and the lightest green representing a Shore value of 0. (b) Comparison showing the effects of grain size and shape on shock absorbance for 3D printed grains.}
\end{figure}

Results can be seen in Fig.~\ref{fig:shockresults}, and are summarised in Table~\ref{tab:results_summary}.  Statistical significance is assessed via a Mann-Whitney U-test with a p-value $<0.05$ and is provided in Appendix Table~\ref{tab:shock_MW}.

\subsection{Size Effects}

Rigid grain performance is dominated by size effects (Fig.~\ref{fig:shockresults}(b)).  3D printed grains show a monolithic increase in shock absorbance with grain size, with increased damping performance observed as size increases. Results show larger grains (7 SVE) significantly outperform smaller variations (3 SVE) for the same shape/material combination in printed grains.  Visually, larger grains appeared to spread upon impact whereas the smaller grains behaved closer to a solid. Grain material has a significant impact on this result, as discussed later.

\subsection{Shape Effects}

Shape effects were more readily seen in some grain sizes, e.g. superellipsoids (again an unconventional shape, not frequently studied in the literature) outperforming spheres at 3mm, 4mm, and 7mm SVE. This is likely due to the tendency of superellipsoids to pack less densely than spheres, allowing more 'give' during impact as grains can move more freely to absorb forces.  This requires further evaluation, as simulation studies on sheared granular materials indicate that damping ratio decreases with increasing aspect ratio ~\cite{tong2015simulations}.  Our other grain shapes didn't show any notable strong trends.

\subsection{Material Effects}

Two interesting material effects are observed.  First, and contrary to grip testing, soft grains perform worse than their rigid counterparts.  Rubber crumb and Agilus grains performed particularly poorly due to their tendency to bounce upon impact, further discouraging their use. Similar to gripping, long-term effects of compressing soft grains are open to further investigation. The rigid Vero, solid plastic, and glass grains were found to be the best materials for shock absorbance. 

Second, the increase in performance with increasing grain size in printed grains is not observed for other (non printed) grains. The material and material deposition technique likely play a key role in this trend, and further investigation with a range of printer settings is required.

\begin{table}[ht]
\caption{Results Summary: Shock absorption (G's) and retention force (N) average and (standard error) for each grain type}
\label{tab:results_summary}
\resizebox{\columnwidth}{!}{%
\begin{tabular}{cccccc}
\hline
\textbf{\begin{tabular}[c]{@{}c@{}}Grain\\ Code\end{tabular}} & \textbf{\begin{tabular}[c]{@{}c@{}}Shock\\ Absorption\end{tabular}} & \textbf{\begin{tabular}[c]{@{}c@{}}Ball\\ Retention\end{tabular}} & \textbf{\begin{tabular}[c]{@{}c@{}}Coin\\ Retention\end{tabular}} & \textbf{\begin{tabular}[c]{@{}c@{}}Cube\\ Retention\end{tabular}} & \textbf{\begin{tabular}[c]{@{}c@{}}Star\\ Retention\end{tabular}} \\ \hline
\textbf{E 3SVE V} & 8.06 (0.38) & 1.12 (0.19) & 0.01 (0.01) & 1.74 (0.29) & 1.63 (0.54) \\
\textbf{SE 3SVE V} & 6.99 (0.61) & 1.23 (0.44) & 0.11 (0.06) & 1.34 (0.13) & 1.67 (0.45) \\
\textbf{C 3SVE V} & 6.5 (0.16) & 3.96 (0.58) & 1.24 (0.28) & 3.44 (0.47) & 4.06 (0.72) \\
\textbf{S 3SVE V} & 7.84 (0.21) & 2.43 (0.7) & 0.39 (0.11) & 2.53 (0.32) & 1.75 (0.6) \\
\textbf{E 4SVE V} & 8.06 (0.41) & 0.39 (0.1) & 0.12 (0.06) & 0.87 (0.17) & 0.8 (0.27) \\
\textbf{SE 4SVE V} & 6.4 (0.29) & 0.11 (0.06) & 0.01 (0.01) & 1.48 (0.42) & 1.16 (0.31) \\
\textbf{S 4VE V} & 6.86 (0.07) & 0.83 (0.28) & 0.43 (0.15) & 1.46 (0.34) & 0.92 (0.23) \\
\textbf{C 4SVE V} & 5.32 (0.42) & 1.1 (0.27) & 0.01 (0.01) & 3.69 (0.64) & 3.42 (0.58) \\
\textbf{E 5SVE V} & 7.78 (0.29) & 1.85 (0.32) & 0.2 (0.12) & 2.23 (0.33) & 5.01 (1.06) \\
\textbf{SE 5SVE V} & 5.96 (0.17) & 1.23 (0.34) & 0.06 (0.02) & 1.7 (0.14) & 1.66 (0.39) \\
\textbf{C 5SVE V} & 5.31 (0.32) & 1.77 (0.44) & 0.02 (0.01) & 3.27 (0.47) & 4.32 (1.08) \\
\textbf{S 5SVE V} & 5.63 (0.18) & 1.41 (0.36) & 0.14 (0.05) & 1.66 (0.34) & 2.07 (0.51) \\
\textbf{E 6SVE V} & 6.36 (0.37) & 1.23 (0.35) & 0.03 (0.02) & 2.5 (0.45) & 2.4 (0.51) \\
\textbf{SE 6SVE V} & 6.2 (0.11) & 1.11 (0.39) & 0.09 (0.07) & 2.8 (0.55) & 3.88 (0.87) \\
\textbf{C 6SVE V} & 5.49 (0.28) & 2.54 (0.67) & 0.18 (0.1) & 2.91 (0.68) & 5.64 (1.38) \\
\textbf{S 6SVE V} & 5.18 (0.24) & 1.33 (0.47) & 0.04 (0.03) & 1.9 (0.63) & 2.53 (0.65) \\
\textbf{E 7SVE V} & 5.73 (0.26) & 1.5 (0.38) & 0.03 (0.02) & 2.27 (0.73) & 3.58 (0.79) \\
\textbf{SE 7SVE V} & 4.16 (0.15) & 1.24 (0.39) & 0.07 (0.04) & 5.01 (0.69) & 7.77 (0.78) \\
\textbf{C 7SVE V} & 4.85 (0.24) & 0.79 (0.18) & 0.02 (0.01) & 4.18 (0.52) & 7.44 (1.17) \\
\textbf{S 7SVE V} & 4.55 (0.21) & 1.31 (0.36) & 0.02 (0.01) & 2.69 (0.5) & 5.41 (1.4) \\
\textbf{S 3 G} & 6.13 (0.15) & 0.05 (0.02) & 0.01 (0.01) & 1.03 (0.12) & 1.1 (0.31) \\
\textbf{S 6 G} & 4.74 (0.16) & 0.21 (0.09) & 0.04 (0.01) & 1.93 (0.41) & 2.05 (0.47) \\
\textbf{S 8 G} & 2.08 (0.07) & 0.83 (0.29) & 0.02 (0.01) & 1.57 (0.44) & 2.58 (1.08) \\
\textbf{S 3 SP} & 5.01 (0.32) & 0.68 (0.14) & 0.14 (0.07) & 1.4 (0.37) & 2.9 (0.46) \\
\textbf{C 7 HP} & 2.9 (0.11) & 2.59 (0.38) & 0.05 (0.04) & 5.45 (0.53) & 9.95 (2.19) \\
\textbf{C 3 HP} & 7.42 (0.09) & 0.94 (0.26) & 0.02 (0.02) & 2.77 (0.46) & 2.91 (0.68) \\
\textbf{S 7 HP} & 4.78 (0.26) & 0.52 (0.2) & 0.01 (0.01) & 2.31 (0.55) & 2.95 (1.12) \\
\textbf{Rubber} & 8.52 (0.21) & 8.9 (1.13) & 0.84 (0.18) & 9.36 (0.64) & 13.02 (1.12) \\
\textbf{Coffee} & 4.46 (0.1) & 3.9 (0.27) & 1.00 (0.25) & 4.53 (0.23) & 2.29 (0.39) \\
\textbf{Rice} & - & 0.3 (0.07) & 0.44 (0.15) & 1.06 (0.19) & 3.07 (0.5) \\
\textbf{Polystyrene} & - & 7.82 (0.57) & 1.07 (0.22) & 8.84 (0.37) & 4.58 (0.48) \\
\textbf{SEBS0} & 8.28 (0.16) & 2.4 (0.29) & 1.3 (0.18) & 5.91 (0.43) & 1.38 (0.26) \\
\textbf{SEBS30} & 6.51 (0.20) & 0.9 (0.06) & 0.6 (0.15) & 2.46 (0.13) & 0.72 (0.18) \\
\textbf{C 4 AG30} & - & 8.62 (0.87) & 1.55 (0.28) & 6.75 (0.68) & 3.86 (0.83) \\
\textbf{C 4 AG70} & - & 5.47 (0.55) & 1.02 (0.38) & 12.66 (0.69) & 13.44 (0.94) \\ %
\end{tabular}%
}
\end{table}

\section{Discussion}
We have presented the largest and most diverse data set of grain types for granular jamming applications in soft robotics, quantifying both gripping performance and shock absorbance.  Manufactured, natural, and printed grains are all represented in the data set.  Key takeaways include the overall higher performance of soft grains for gripping, and large, rigid grains for shock absorption. The data contained presented here offers an exciting avenue for the principled creation and optimisation of more efficient jamming designs, where specific grains can be selected based on the performance requirements for a given soft robotic system. 

Our experimental setups are designed to provide accurate results from readily available and cheap components.  Moreover, CAD files for printed components are available publicly at the data set URL provided in the Abstract.  This decision was made to allow researchers to confirm our findings, and experiment with other grain types whilst having some baseline comparisons to refer to from our data set.  Researchers can also contribute their data to this data set, or suggest grains for us to add, and we encourage interested researchers to reach out to do so.  In future work we aim to release additional sets of grains to further the coverage and utility of the data set, as well as delve more deeply into trends uncovered in this experimentation, e.g., the effect of printer settings on printed grain properties.

\bibliographystyle{IEEEtran}
\bibliography{main}

 \appendix
 
Mann-Whitney U Test results for grip and shock absorption tests.  No colour = no significance.  Blue = row is statistically superior to column. Orange = column is statistically superior to row.  Statistical significance assessed at P<0.05.
 
\begin{table}[ht]
\caption{Cube Retention Mann-Whitney Row/Col comparison}
\label{tab:cube_grip}
\resizebox{\columnwidth}{!}{%
\begin{tabular}{|c|llllllllllllllllllllllllllllllllll|}
\hline
\rowcolor[HTML]{D9E1F2}
\cellcolor[HTML]{FFFFFF} & \rotatebox[origin=c]{90}{C 3SVE V} & \rotatebox[origin=c]{90}{E 3SVE V} & \rotatebox[origin=c]{90}{S 3SVE V} & \rotatebox[origin=c]{90}{SE 3SVE V} & \rotatebox[origin=c]{90}{C 4SVE V} & \rotatebox[origin=c]{90}{E 4SVE V} & \rotatebox[origin=c]{90}{S 4SVE V} & \rotatebox[origin=c]{90}{SE 4SVE V} & \rotatebox[origin=c]{90}{C 5SVE V} & \rotatebox[origin=c]{90}{E 5SVE V} & \rotatebox[origin=c]{90}{S 5SVE V} & \rotatebox[origin=c]{90}{SE 5SVE V} & \rotatebox[origin=c]{90}{C 6SVE V} & \rotatebox[origin=c]{90}{E 6SVE V} & \rotatebox[origin=c]{90}{S 6SVE V} & \rotatebox[origin=c]{90}{SE 6SVE V} & \rotatebox[origin=c]{90}{C 7SVE V} & \rotatebox[origin=c]{90}{E 7SVE V} & \rotatebox[origin=c]{90}{S 7SVE V} & \rotatebox[origin=c]{90}{SE 7SVE V} & \rotatebox[origin=c]{90}{S 3 G} & \rotatebox[origin=c]{90}{S 6 G} & \rotatebox[origin=c]{90}{S 8 G} & \rotatebox[origin=c]{90}{S 3 SP} & \rotatebox[origin=c]{90}{C 3 HP} & \rotatebox[origin=c]{90}{C 7 HP} & \rotatebox[origin=c]{90}{S 7 HP} & \rotatebox[origin=c]{90}{Rice} & \rotatebox[origin=c]{90}{Coffee} & \rotatebox[origin=c]{90}{Polystyrene} & \rotatebox[origin=c]{90}{Rubber} & \rotatebox[origin=c]{90}{C 4 AG70} & \rotatebox[origin=c]{90}{C 4 AG30} & \rotatebox[origin=c]{90}{SEBS30} \\ \hline
\rowcolor[HTML]{8EA9DB} 
\cellcolor[HTML]{FCE4D6}SEBS0 &  &  &  &  &  &  &  &  &  &  &  &  &  &  &  &  &  &  &  & \cellcolor[HTML]{F2F2F2} &  &  &  &  &  & \cellcolor[HTML]{F2F2F2} &  &  &  & \cellcolor[HTML]{F4B084} & \cellcolor[HTML]{F4B084} & \cellcolor[HTML]{F4B084} & \cellcolor[HTML]{F2F2F2} &  \\
\cellcolor[HTML]{FCE4D6}SEBS30 & \cellcolor[HTML]{F2F2F2} & \cellcolor[HTML]{8EA9DB} & \cellcolor[HTML]{F2F2F2} & \cellcolor[HTML]{8EA9DB} & \cellcolor[HTML]{F2F2F2} & \cellcolor[HTML]{F4B084} & \cellcolor[HTML]{8EA9DB} & \cellcolor[HTML]{8EA9DB} & \cellcolor[HTML]{F2F2F2} & \cellcolor[HTML]{F2F2F2} & \cellcolor[HTML]{8EA9DB} & \cellcolor[HTML]{8EA9DB} & \cellcolor[HTML]{F2F2F2} & \cellcolor[HTML]{F2F2F2} & \cellcolor[HTML]{8EA9DB} & \cellcolor[HTML]{F2F2F2} & \cellcolor[HTML]{F4B084} & \cellcolor[HTML]{F2F2F2} & \cellcolor[HTML]{F2F2F2} & \cellcolor[HTML]{F4B084} & \cellcolor[HTML]{8EA9DB} & \cellcolor[HTML]{F2F2F2} & \cellcolor[HTML]{8EA9DB} & \cellcolor[HTML]{8EA9DB} & \cellcolor[HTML]{F2F2F2} & \cellcolor[HTML]{F4B084} & \cellcolor[HTML]{F2F2F2} & \cellcolor[HTML]{F4B084} & \cellcolor[HTML]{F4B084} & \cellcolor[HTML]{F4B084} & \cellcolor[HTML]{F4B084} & \cellcolor[HTML]{F4B084} & \cellcolor[HTML]{F4B084} & \cellcolor[HTML]{BFBFBF} \\
\rowcolor[HTML]{8EA9DB} 
\cellcolor[HTML]{FCE4D6}C 4 AG30 &  &  &  &  &  &  &  &  &  &  &  &  &  &  &  &  &  &  &  & \cellcolor[HTML]{F2F2F2} &  &  &  &  &  & \cellcolor[HTML]{F2F2F2} &  &  &  & \cellcolor[HTML]{F4B084} & \cellcolor[HTML]{F4B084} & \cellcolor[HTML]{F4B084} & \cellcolor[HTML]{BFBFBF} & \cellcolor[HTML]{F2F2F2} \\
\rowcolor[HTML]{8EA9DB} 
\cellcolor[HTML]{FCE4D6}C 4 AG70 &  &  &  &  &  &  &  &  &  &  &  &  &  &  &  &  &  &  &  &  &  &  &  &  &  &  &  &  &  &  &  & \cellcolor[HTML]{BFBFBF} & \cellcolor[HTML]{F2F2F2} & \cellcolor[HTML]{F2F2F2} \\
\rowcolor[HTML]{8EA9DB} 
\cellcolor[HTML]{FCE4D6}Rubber &  &  &  &  &  &  &  &  &  &  &  &  &  &  &  &  &  &  &  &  &  &  &  &  &  &  &  &  &  & \cellcolor[HTML]{F2F2F2} & \cellcolor[HTML]{BFBFBF} & \cellcolor[HTML]{F2F2F2} & \cellcolor[HTML]{F2F2F2} & \cellcolor[HTML]{F2F2F2} \\
\rowcolor[HTML]{8EA9DB} 
\cellcolor[HTML]{FCE4D6}Polystyrene &  &  &  &  &  &  &  &  &  &  &  &  &  &  &  &  &  &  &  &  &  &  &  &  &  &  &  &  &  & \cellcolor[HTML]{BFBFBF} & \cellcolor[HTML]{F2F2F2} & \cellcolor[HTML]{F2F2F2} & \cellcolor[HTML]{F2F2F2} & \cellcolor[HTML]{F2F2F2} \\
\rowcolor[HTML]{8EA9DB} 
\cellcolor[HTML]{FCE4D6}Coffee & \cellcolor[HTML]{F2F2F2} &  &  &  & \cellcolor[HTML]{F2F2F2} &  &  &  &  &  &  &  & \cellcolor[HTML]{F2F2F2} &  &  &  & \cellcolor[HTML]{F2F2F2} &  &  & \cellcolor[HTML]{F2F2F2} &  &  &  &  &  & \cellcolor[HTML]{F2F2F2} &  &  & \cellcolor[HTML]{BFBFBF} & \cellcolor[HTML]{F2F2F2} & \cellcolor[HTML]{F2F2F2} & \cellcolor[HTML]{F2F2F2} & \cellcolor[HTML]{F2F2F2} & \cellcolor[HTML]{F2F2F2} \\
\rowcolor[HTML]{F2F2F2} 
\cellcolor[HTML]{FCE4D6}Rice & \cellcolor[HTML]{8EA9DB} & \cellcolor[HTML]{8EA9DB} & \cellcolor[HTML]{8EA9DB} &  & \cellcolor[HTML]{8EA9DB} &  &  &  & \cellcolor[HTML]{8EA9DB} & \cellcolor[HTML]{8EA9DB} &  & \cellcolor[HTML]{8EA9DB} & \cellcolor[HTML]{8EA9DB} & \cellcolor[HTML]{8EA9DB} &  & \cellcolor[HTML]{8EA9DB} & \cellcolor[HTML]{F4B084} &  & \cellcolor[HTML]{8EA9DB} & \cellcolor[HTML]{F4B084} &  &  &  &  & \cellcolor[HTML]{8EA9DB} & \cellcolor[HTML]{F4B084} &  & \cellcolor[HTML]{BFBFBF} &  &  &  &  &  &  \\
\rowcolor[HTML]{F2F2F2} 
\cellcolor[HTML]{FCE4D6}S 7 HP &  &  &  &  &  & \cellcolor[HTML]{F4B084} &  &  &  &  &  &  &  &  &  &  & \cellcolor[HTML]{F4B084} &  &  & \cellcolor[HTML]{F4B084} &  &  &  &  &  & \cellcolor[HTML]{F4B084} & \cellcolor[HTML]{BFBFBF} &  &  &  &  &  &  &  \\
\rowcolor[HTML]{8EA9DB} 
\cellcolor[HTML]{FCE4D6}C 7 HP &  &  &  &  & \cellcolor[HTML]{F2F2F2} &  &  &  &  &  &  &  &  &  &  &  & \cellcolor[HTML]{F2F2F2} &  &  & \cellcolor[HTML]{F2F2F2} &  &  &  &  &  & \cellcolor[HTML]{BFBFBF} & \cellcolor[HTML]{F2F2F2} & \cellcolor[HTML]{F2F2F2} & \cellcolor[HTML]{F2F2F2} & \cellcolor[HTML]{F2F2F2} & \cellcolor[HTML]{F2F2F2} & \cellcolor[HTML]{F2F2F2} & \cellcolor[HTML]{F2F2F2} & \cellcolor[HTML]{F2F2F2} \\
\rowcolor[HTML]{F2F2F2} 
\cellcolor[HTML]{FCE4D6}C 3 HP &  &  &  & \cellcolor[HTML]{F4B084} &  & \cellcolor[HTML]{F4B084} & \cellcolor[HTML]{8EA9DB} & \cellcolor[HTML]{8EA9DB} &  &  & \cellcolor[HTML]{8EA9DB} & \cellcolor[HTML]{8EA9DB} &  &  &  &  &  &  &  & \cellcolor[HTML]{F4B084} & \cellcolor[HTML]{F4B084} &  & \cellcolor[HTML]{8EA9DB} & \cellcolor[HTML]{8EA9DB} & \cellcolor[HTML]{BFBFBF} &  &  &  &  &  &  &  &  &  \\
\rowcolor[HTML]{F2F2F2} 
\cellcolor[HTML]{FCE4D6}S 3 SP & \cellcolor[HTML]{F4B084} &  & \cellcolor[HTML]{8EA9DB} &  & \cellcolor[HTML]{F4B084} &  &  &  & \cellcolor[HTML]{F4B084} &  &  &  & \cellcolor[HTML]{8EA9DB} &  &  & \cellcolor[HTML]{8EA9DB} & \cellcolor[HTML]{F4B084} &  &  & \cellcolor[HTML]{F4B084} &  &  &  & \cellcolor[HTML]{BFBFBF} &  &  &  &  &  &  &  &  &  &  \\
\rowcolor[HTML]{F2F2F2} 
\cellcolor[HTML]{FCE4D6}S 8 G & \cellcolor[HTML]{F4B084} &  &  &  & \cellcolor[HTML]{F4B084} &  &  &  & \cellcolor[HTML]{F4B084} &  &  &  &  &  &  &  & \cellcolor[HTML]{F4B084} &  &  & \cellcolor[HTML]{F4B084} &  &  & \cellcolor[HTML]{BFBFBF} &  &  &  &  &  &  &  &  &  &  &  \\
\rowcolor[HTML]{F2F2F2} 
\cellcolor[HTML]{FCE4D6}S 6 G & \cellcolor[HTML]{F4B084} &  &  &  & \cellcolor[HTML]{F4B084} &  &  &  &  &  &  &  &  &  &  &  & \cellcolor[HTML]{F4B084} &  &  & \cellcolor[HTML]{F4B084} &  & \cellcolor[HTML]{BFBFBF} &  &  &  &  &  &  &  &  &  &  &  &  \\
\rowcolor[HTML]{F2F2F2} 
\cellcolor[HTML]{FCE4D6}S 3 G & \cellcolor[HTML]{8EA9DB} &  & \cellcolor[HTML]{8EA9DB} &  & \cellcolor[HTML]{8EA9DB} &  &  &  & \cellcolor[HTML]{8EA9DB} & \cellcolor[HTML]{8EA9DB} &  & \cellcolor[HTML]{8EA9DB} & \cellcolor[HTML]{8EA9DB} & \cellcolor[HTML]{8EA9DB} &  & \cellcolor[HTML]{8EA9DB} & \cellcolor[HTML]{F4B084} &  & \cellcolor[HTML]{8EA9DB} & \cellcolor[HTML]{F4B084} & \cellcolor[HTML]{BFBFBF} &  &  &  &  &  &  &  &  &  &  &  &  &  \\
\rowcolor[HTML]{F2F2F2} 
\cellcolor[HTML]{FCE4D6}SE 7SVE V &  & \cellcolor[HTML]{8EA9DB} & \cellcolor[HTML]{8EA9DB} & \cellcolor[HTML]{8EA9DB} &  & \cellcolor[HTML]{8EA9DB} & \cellcolor[HTML]{8EA9DB} & \cellcolor[HTML]{8EA9DB} &  & \cellcolor[HTML]{8EA9DB} & \cellcolor[HTML]{8EA9DB} & \cellcolor[HTML]{8EA9DB} & \cellcolor[HTML]{8EA9DB} & \cellcolor[HTML]{8EA9DB} & \cellcolor[HTML]{8EA9DB} & \cellcolor[HTML]{8EA9DB} &  & \cellcolor[HTML]{8EA9DB} & \cellcolor[HTML]{8EA9DB} & \cellcolor[HTML]{BFBFBF} &  &  &  &  &  &  &  &  &  &  &  &  &  &  \\
\rowcolor[HTML]{F2F2F2} 
\cellcolor[HTML]{FCE4D6}S 7SVE V &  &  &  & \cellcolor[HTML]{F4B084} &  & \cellcolor[HTML]{F4B084} &  & \cellcolor[HTML]{F4B084} &  &  &  &  &  &  &  &  &  &  & \cellcolor[HTML]{BFBFBF} &  &  &  &  &  &  &  &  &  &  &  &  &  &  &  \\
\rowcolor[HTML]{F2F2F2} 
\cellcolor[HTML]{FCE4D6}E 7SVE V &  &  &  &  &  &  &  &  &  &  &  &  &  &  &  &  & \cellcolor[HTML]{F4B084} & \cellcolor[HTML]{BFBFBF} &  &  &  &  &  &  &  &  &  &  &  &  &  &  &  &  \\
\rowcolor[HTML]{F2F2F2} 
\cellcolor[HTML]{FCE4D6}C 7SVE V &  & \cellcolor[HTML]{8EA9DB} & \cellcolor[HTML]{8EA9DB} & \cellcolor[HTML]{8EA9DB} &  & \cellcolor[HTML]{8EA9DB} & \cellcolor[HTML]{8EA9DB} & \cellcolor[HTML]{8EA9DB} &  & \cellcolor[HTML]{8EA9DB} & \cellcolor[HTML]{8EA9DB} & \cellcolor[HTML]{8EA9DB} &  & \cellcolor[HTML]{8EA9DB} & \cellcolor[HTML]{8EA9DB} &  & \cellcolor[HTML]{BFBFBF} &  &  &  &  &  &  &  &  &  &  &  &  &  &  &  &  &  \\
\rowcolor[HTML]{F2F2F2} 
\cellcolor[HTML]{FCE4D6}SE 6SVE V &  &  &  & \cellcolor[HTML]{F4B084} &  & \cellcolor[HTML]{F4B084} &  & \cellcolor[HTML]{F4B084} &  &  &  &  &  &  &  & \cellcolor[HTML]{BFBFBF} &  &  &  &  &  &  &  &  &  &  &  &  &  &  &  &  &  &  \\
\rowcolor[HTML]{F2F2F2} 
\cellcolor[HTML]{FCE4D6}S 6SVE V & \cellcolor[HTML]{F4B084} &  &  &  & \cellcolor[HTML]{F4B084} &  &  &  & \cellcolor[HTML]{F4B084} &  &  &  &  &  & \cellcolor[HTML]{BFBFBF} &  &  &  &  &  &  &  &  &  &  &  &  &  &  &  &  &  &  &  \\
\rowcolor[HTML]{F2F2F2} 
\cellcolor[HTML]{FCE4D6}E 6SVE V &  &  &  & \cellcolor[HTML]{F4B084} &  & \cellcolor[HTML]{F4B084} &  &  &  &  &  &  &  & \cellcolor[HTML]{BFBFBF} &  &  &  &  &  &  &  &  &  &  &  &  &  &  &  &  &  &  &  &  \\
\rowcolor[HTML]{F2F2F2} 
\cellcolor[HTML]{FCE4D6}C 6SVE V &  &  &  & \cellcolor[HTML]{F4B084} &  & \cellcolor[HTML]{F4B084} &  & \cellcolor[HTML]{F4B084} &  &  & \cellcolor[HTML]{F4B084} &  & \cellcolor[HTML]{BFBFBF} &  &  &  &  &  &  &  &  &  &  &  &  &  &  &  &  &  &  &  &  &  \\
\rowcolor[HTML]{F2F2F2} 
\cellcolor[HTML]{FCE4D6}SE 5SVE V & \cellcolor[HTML]{F4B084} &  &  &  & \cellcolor[HTML]{F4B084} & \cellcolor[HTML]{F4B084} &  &  & \cellcolor[HTML]{F4B084} &  &  & \cellcolor[HTML]{BFBFBF} &  &  &  &  &  &  &  &  &  &  &  &  &  &  &  &  &  &  &  &  &  &  \\
\rowcolor[HTML]{F2F2F2} 
\cellcolor[HTML]{FCE4D6}S 5SVE V & \cellcolor[HTML]{F4B084} &  &  &  & \cellcolor[HTML]{F4B084} &  &  &  & \cellcolor[HTML]{F4B084} &  & \cellcolor[HTML]{BFBFBF} &  &  &  &  &  &  &  &  &  &  &  &  &  &  &  &  &  &  &  &  &  &  &  \\
\rowcolor[HTML]{F2F2F2} 
\cellcolor[HTML]{FCE4D6}E 5SVE V &  &  &  &  & \cellcolor[HTML]{F4B084} & \cellcolor[HTML]{F4B084} &  &  &  & \cellcolor[HTML]{BFBFBF} &  &  &  &  &  &  &  &  &  &  &  &  &  &  &  &  &  &  &  &  &  &  &  &  \\
\rowcolor[HTML]{F2F2F2} 
\cellcolor[HTML]{FCE4D6}C 5SVE V &  & \cellcolor[HTML]{8EA9DB} &  & \cellcolor[HTML]{8EA9DB} &  & \cellcolor[HTML]{F4B084} & \cellcolor[HTML]{8EA9DB} & \cellcolor[HTML]{8EA9DB} & \cellcolor[HTML]{BFBFBF} &  &  &  &  &  &  &  &  &  &  &  &  &  &  &  &  &  &  &  &  &  &  &  &  &  \\
\rowcolor[HTML]{F2F2F2} 
\cellcolor[HTML]{FCE4D6}SE 4SVE V & \cellcolor[HTML]{F4B084} &  &  &  & \cellcolor[HTML]{F4B084} &  &  & \cellcolor[HTML]{BFBFBF} &  &  &  &  &  &  &  &  &  &  &  &  &  &  &  &  &  &  &  &  &  &  &  &  &  &  \\
\rowcolor[HTML]{F2F2F2} 
\cellcolor[HTML]{FCE4D6}S 4SVE V & \cellcolor[HTML]{F4B084} &  &  &  & \cellcolor[HTML]{F4B084} &  & \cellcolor[HTML]{BFBFBF} &  &  &  &  &  &  &  &  &  &  &  &  &  &  &  &  &  &  &  &  &  &  &  &  &  &  &  \\
\rowcolor[HTML]{F2F2F2} 
\cellcolor[HTML]{FCE4D6}E 4SVE V & \cellcolor[HTML]{8EA9DB} & \cellcolor[HTML]{8EA9DB} & \cellcolor[HTML]{8EA9DB} &  & \cellcolor[HTML]{8EA9DB} & \cellcolor[HTML]{BFBFBF} &  &  &  &  &  &  &  &  &  &  &  &  &  &  &  &  &  &  &  &  &  &  &  &  &  &  &  &  \\
\rowcolor[HTML]{F2F2F2} 
\cellcolor[HTML]{FCE4D6}C 4SVE V &  & \cellcolor[HTML]{8EA9DB} &  & \cellcolor[HTML]{8EA9DB} & \cellcolor[HTML]{BFBFBF} &  &  &  &  &  &  &  &  &  &  &  &  &  &  &  &  &  &  &  &  &  &  &  &  &  &  &  &  &  \\
\rowcolor[HTML]{F2F2F2} 
\cellcolor[HTML]{FCE4D6}SE 3SVE V & \cellcolor[HTML]{8EA9DB} &  & \cellcolor[HTML]{8EA9DB} & \cellcolor[HTML]{BFBFBF} &  &  &  &  &  &  &  &  &  &  &  &  &  &  &  &  &  &  &  &  &  &  &  &  &  &  &  &  &  &  \\
\rowcolor[HTML]{F2F2F2} 
\cellcolor[HTML]{FCE4D6}S 3SVE V &  &  & \cellcolor[HTML]{BFBFBF} &  &  &  &  &  &  &  &  &  &  &  &  &  &  &  &  &  &  &  &  &  &  &  &  &  &  &  &  &  &  &  \\
\rowcolor[HTML]{F2F2F2} 
\cellcolor[HTML]{FCE4D6}E 3SVE V & \cellcolor[HTML]{F4B084} & \cellcolor[HTML]{BFBFBF} &  &  &  &  &  &  &  &  &  &  &  &  &  &  &  &  &  &  &  &  &  &  &  &  &  &  &  &  &  &  &  &  \\ \hline
\end{tabular}%
    }
\end{table}

\begin{table}[ht]
\caption{Star Retention Mann-Whitney Row/Col comparison}
\label{tab:star_grip}
\resizebox{\columnwidth}{!}{%
\begin{tabular}{|c|llllllllllllllllllllllllllllllllll|}
\hline
\rowcolor[HTML]{D9E1F2}
\cellcolor[HTML]{FFFFFF} & \rotatebox[origin=c]{90}{C 3SVE V} & \rotatebox[origin=c]{90}{E 3SVE V} & \rotatebox[origin=c]{90}{S 3SVE V} & \rotatebox[origin=c]{90}{SE 3SVE V} & \rotatebox[origin=c]{90}{C 4SVE V} & \rotatebox[origin=c]{90}{E 4SVE V} & \rotatebox[origin=c]{90}{S 4SVE V} & \rotatebox[origin=c]{90}{SE 4SVE V} & \rotatebox[origin=c]{90}{C 5SVE V} & \rotatebox[origin=c]{90}{E 5SVE V} & \rotatebox[origin=c]{90}{S 5SVE V} & \rotatebox[origin=c]{90}{SE 5SVE V} & \rotatebox[origin=c]{90}{C 6SVE V} & \rotatebox[origin=c]{90}{E 6SVE V} & \rotatebox[origin=c]{90}{S 6SVE V} & \rotatebox[origin=c]{90}{SE 6SVE V} & \rotatebox[origin=c]{90}{C 7SVE V} & \rotatebox[origin=c]{90}{E 7SVE V} & \rotatebox[origin=c]{90}{S 7SVE V} & \rotatebox[origin=c]{90}{SE 7SVE V} & \rotatebox[origin=c]{90}{S 3 G} & \rotatebox[origin=c]{90}{S 6 G} & \rotatebox[origin=c]{90}{S 8 G} & \rotatebox[origin=c]{90}{S 3 SP} & \rotatebox[origin=c]{90}{C 3 HP} & \rotatebox[origin=c]{90}{C 7 HP} & \rotatebox[origin=c]{90}{S 7 HP} & \rotatebox[origin=c]{90}{Rice} & \rotatebox[origin=c]{90}{Coffee} & \rotatebox[origin=c]{90}{Polystyrene} & \rotatebox[origin=c]{90}{Rubber} & \rotatebox[origin=c]{90}{C 4 AG70} & \rotatebox[origin=c]{90}{C 4 AG30} & \rotatebox[origin=c]{90}{SEBS30} \\ \hline
\rowcolor[HTML]{F2F2F2} 
\cellcolor[HTML]{FCE4D6}SEBS0 & \cellcolor[HTML]{8EA9DB} &  &  &  & \cellcolor[HTML]{8EA9DB} &  &  &  & \cellcolor[HTML]{8EA9DB} & \cellcolor[HTML]{8EA9DB} &  &  & \cellcolor[HTML]{8EA9DB} &  &  &  & \cellcolor[HTML]{F4B084} & \cellcolor[HTML]{8EA9DB} & \cellcolor[HTML]{8EA9DB} & \cellcolor[HTML]{F4B084} &  &  & \cellcolor[HTML]{8EA9DB} &  & \cellcolor[HTML]{F4B084} &  &  &  &  & \cellcolor[HTML]{F4B084} & \cellcolor[HTML]{F4B084} & \cellcolor[HTML]{F4B084} & \cellcolor[HTML]{8EA9DB} & \cellcolor[HTML]{F4B084} \\
\rowcolor[HTML]{8EA9DB} 
\cellcolor[HTML]{FCE4D6}SEBS30 &  & \cellcolor[HTML]{F2F2F2} &  & \cellcolor[HTML]{F2F2F2} &  & \cellcolor[HTML]{F2F2F2} & \cellcolor[HTML]{F2F2F2} & \cellcolor[HTML]{F2F2F2} &  &  &  &  &  & \cellcolor[HTML]{F2F2F2} &  &  & \cellcolor[HTML]{F4B084} &  &  & \cellcolor[HTML]{F4B084} & \cellcolor[HTML]{F2F2F2} &  &  &  & \cellcolor[HTML]{F4B084} & \cellcolor[HTML]{F2F2F2} & \cellcolor[HTML]{F2F2F2} &  & \cellcolor[HTML]{F2F2F2} &  & \cellcolor[HTML]{F4B084} & \cellcolor[HTML]{F4B084} &  & \cellcolor[HTML]{BFBFBF} \\
\rowcolor[HTML]{F2F2F2} 
\cellcolor[HTML]{FCE4D6}C 4 AG30 &  & \cellcolor[HTML]{8EA9DB} & \cellcolor[HTML]{8EA9DB} & \cellcolor[HTML]{8EA9DB} &  & \cellcolor[HTML]{F4B084} & \cellcolor[HTML]{F4B084} & \cellcolor[HTML]{F4B084} &  &  & \cellcolor[HTML]{8EA9DB} & \cellcolor[HTML]{8EA9DB} &  &  &  &  & \cellcolor[HTML]{F4B084} &  &  & \cellcolor[HTML]{F4B084} & \cellcolor[HTML]{8EA9DB} &  &  &  & \cellcolor[HTML]{F4B084} &  &  &  &  &  & \cellcolor[HTML]{F4B084} & \cellcolor[HTML]{F4B084} & \cellcolor[HTML]{BFBFBF} &  \\
\rowcolor[HTML]{8EA9DB} 
\cellcolor[HTML]{FCE4D6}C 4 AG70 &  &  &  &  &  &  &  &  &  &  &  &  &  &  &  &  &  &  &  &  &  &  &  &  & \cellcolor[HTML]{F2F2F2} &  &  &  &  &  & \cellcolor[HTML]{F2F2F2} & \cellcolor[HTML]{BFBFBF} & \cellcolor[HTML]{F2F2F2} & \cellcolor[HTML]{F2F2F2} \\
\rowcolor[HTML]{8EA9DB} 
\cellcolor[HTML]{FCE4D6}Rubber &  &  &  &  &  &  &  &  &  &  &  &  &  &  &  &  &  &  &  &  &  &  &  &  & \cellcolor[HTML]{F2F2F2} &  &  &  &  &  & \cellcolor[HTML]{BFBFBF} & \cellcolor[HTML]{F2F2F2} & \cellcolor[HTML]{F2F2F2} & \cellcolor[HTML]{F2F2F2} \\
\cellcolor[HTML]{FCE4D6}Polystyrene & \cellcolor[HTML]{F2F2F2} & \cellcolor[HTML]{8EA9DB} & \cellcolor[HTML]{8EA9DB} & \cellcolor[HTML]{8EA9DB} & \cellcolor[HTML]{F2F2F2} & \cellcolor[HTML]{F4B084} & \cellcolor[HTML]{8EA9DB} & \cellcolor[HTML]{8EA9DB} & \cellcolor[HTML]{F2F2F2} & \cellcolor[HTML]{F2F2F2} & \cellcolor[HTML]{8EA9DB} & \cellcolor[HTML]{8EA9DB} & \cellcolor[HTML]{F2F2F2} & \cellcolor[HTML]{8EA9DB} & \cellcolor[HTML]{8EA9DB} & \cellcolor[HTML]{F2F2F2} & \cellcolor[HTML]{F2F2F2} & \cellcolor[HTML]{F2F2F2} & \cellcolor[HTML]{F2F2F2} & \cellcolor[HTML]{F4B084} & \cellcolor[HTML]{8EA9DB} & \cellcolor[HTML]{8EA9DB} & \cellcolor[HTML]{8EA9DB} & \cellcolor[HTML]{F2F2F2} & \cellcolor[HTML]{F4B084} & \cellcolor[HTML]{F2F2F2} & \cellcolor[HTML]{8EA9DB} & \cellcolor[HTML]{F2F2F2} & \cellcolor[HTML]{8EA9DB} & \cellcolor[HTML]{BFBFBF} & \cellcolor[HTML]{F2F2F2} & \cellcolor[HTML]{F2F2F2} & \cellcolor[HTML]{F2F2F2} & \cellcolor[HTML]{F2F2F2} \\
\rowcolor[HTML]{F2F2F2} 
\cellcolor[HTML]{FCE4D6}Coffee &  &  &  &  &  & \cellcolor[HTML]{F4B084} &  &  &  &  &  &  &  &  &  &  & \cellcolor[HTML]{F4B084} &  &  & \cellcolor[HTML]{F4B084} &  &  &  &  & \cellcolor[HTML]{F4B084} &  &  &  & \cellcolor[HTML]{BFBFBF} &  &  &  &  &  \\
\rowcolor[HTML]{F2F2F2} 
\cellcolor[HTML]{FCE4D6}Rice &  &  &  &  &  & \cellcolor[HTML]{F4B084} & \cellcolor[HTML]{F4B084} & \cellcolor[HTML]{F4B084} &  &  &  &  &  &  &  &  & \cellcolor[HTML]{F4B084} &  &  & \cellcolor[HTML]{F4B084} &  &  &  &  & \cellcolor[HTML]{F4B084} &  &  & \cellcolor[HTML]{BFBFBF} &  &  &  &  &  &  \\
\rowcolor[HTML]{F2F2F2} 
\cellcolor[HTML]{FCE4D6}S 7 HP &  &  &  &  &  &  &  &  &  &  &  &  &  &  &  &  & \cellcolor[HTML]{F4B084} &  &  & \cellcolor[HTML]{F4B084} &  &  &  &  & \cellcolor[HTML]{F4B084} &  & \cellcolor[HTML]{BFBFBF} &  &  &  &  &  &  &  \\
\rowcolor[HTML]{F2F2F2} 
\cellcolor[HTML]{FCE4D6}C 7 HP &  &  &  &  &  &  &  &  &  &  &  &  &  &  &  &  & \cellcolor[HTML]{F4B084} &  &  & \cellcolor[HTML]{F4B084} &  &  &  &  & \cellcolor[HTML]{F4B084} & \cellcolor[HTML]{BFBFBF} &  &  &  &  &  &  &  &  \\
\rowcolor[HTML]{8EA9DB} 
\cellcolor[HTML]{FCE4D6}C 3 HP &  &  &  &  &  &  &  &  & \cellcolor[HTML]{F2F2F2} & \cellcolor[HTML]{F2F2F2} &  &  & \cellcolor[HTML]{F2F2F2} &  &  &  & \cellcolor[HTML]{F2F2F2} &  & \cellcolor[HTML]{F2F2F2} & \cellcolor[HTML]{F2F2F2} &  &  &  &  & \cellcolor[HTML]{BFBFBF} & \cellcolor[HTML]{F2F2F2} & \cellcolor[HTML]{F2F2F2} & \cellcolor[HTML]{F2F2F2} & \cellcolor[HTML]{F2F2F2} & \cellcolor[HTML]{F2F2F2} & \cellcolor[HTML]{F2F2F2} & \cellcolor[HTML]{F2F2F2} & \cellcolor[HTML]{F2F2F2} & \cellcolor[HTML]{F2F2F2} \\
\rowcolor[HTML]{F2F2F2} 
\cellcolor[HTML]{FCE4D6}S 3 SP &  &  &  &  &  & \cellcolor[HTML]{F4B084} & \cellcolor[HTML]{F4B084} &  &  &  &  &  &  &  &  &  & \cellcolor[HTML]{F4B084} &  &  & \cellcolor[HTML]{F4B084} &  &  &  & \cellcolor[HTML]{BFBFBF} &  &  &  &  &  &  &  &  &  &  \\
\rowcolor[HTML]{F2F2F2} 
\cellcolor[HTML]{FCE4D6}S 8 G &  &  &  &  &  & \cellcolor[HTML]{F4B084} & \cellcolor[HTML]{F4B084} & \cellcolor[HTML]{F4B084} &  &  &  &  &  &  &  &  & \cellcolor[HTML]{F4B084} &  &  & \cellcolor[HTML]{F4B084} & \cellcolor[HTML]{F4B084} &  & \cellcolor[HTML]{BFBFBF} &  &  &  &  &  &  &  &  &  &  &  \\
\rowcolor[HTML]{F2F2F2} 
\cellcolor[HTML]{FCE4D6}S 6 G &  &  &  &  &  &  &  &  &  & \cellcolor[HTML]{F4B084} &  &  & \cellcolor[HTML]{F4B084} &  &  &  & \cellcolor[HTML]{F4B084} &  & \cellcolor[HTML]{F4B084} & \cellcolor[HTML]{F4B084} &  & \cellcolor[HTML]{BFBFBF} &  &  &  &  &  &  &  &  &  &  &  &  \\
\rowcolor[HTML]{F2F2F2} 
\cellcolor[HTML]{FCE4D6}S 3 G & \cellcolor[HTML]{8EA9DB} &  &  &  & \cellcolor[HTML]{8EA9DB} &  &  &  & \cellcolor[HTML]{8EA9DB} & \cellcolor[HTML]{8EA9DB} &  &  & \cellcolor[HTML]{8EA9DB} &  &  & \cellcolor[HTML]{8EA9DB} & \cellcolor[HTML]{F4B084} & \cellcolor[HTML]{8EA9DB} & \cellcolor[HTML]{8EA9DB} & \cellcolor[HTML]{F4B084} & \cellcolor[HTML]{BFBFBF} &  &  &  &  &  &  &  &  &  &  &  &  &  \\
\rowcolor[HTML]{F2F2F2} 
\cellcolor[HTML]{FCE4D6}SE 7SVE V & \cellcolor[HTML]{8EA9DB} & \cellcolor[HTML]{8EA9DB} & \cellcolor[HTML]{8EA9DB} & \cellcolor[HTML]{8EA9DB} & \cellcolor[HTML]{8EA9DB} & \cellcolor[HTML]{8EA9DB} & \cellcolor[HTML]{8EA9DB} & \cellcolor[HTML]{8EA9DB} & \cellcolor[HTML]{8EA9DB} &  & \cellcolor[HTML]{8EA9DB} & \cellcolor[HTML]{8EA9DB} &  & \cellcolor[HTML]{8EA9DB} & \cellcolor[HTML]{8EA9DB} & \cellcolor[HTML]{8EA9DB} &  & \cellcolor[HTML]{8EA9DB} &  & \cellcolor[HTML]{BFBFBF} &  &  &  &  &  &  &  &  &  &  &  &  &  &  \\
\rowcolor[HTML]{F2F2F2} 
\cellcolor[HTML]{FCE4D6}S 7SVE V &  & \cellcolor[HTML]{8EA9DB} & \cellcolor[HTML]{8EA9DB} & \cellcolor[HTML]{8EA9DB} &  & \cellcolor[HTML]{F4B084} & \cellcolor[HTML]{F4B084} & \cellcolor[HTML]{F4B084} &  &  & \cellcolor[HTML]{8EA9DB} & \cellcolor[HTML]{F4B084} &  &  &  &  &  &  & \cellcolor[HTML]{BFBFBF} &  &  &  &  &  &  &  &  &  &  &  &  &  &  &  \\
\rowcolor[HTML]{F2F2F2} 
\cellcolor[HTML]{FCE4D6}E 7SVE V &  &  &  & \cellcolor[HTML]{F4B084} &  & \cellcolor[HTML]{F4B084} & \cellcolor[HTML]{F4B084} & \cellcolor[HTML]{F4B084} &  &  &  & \cellcolor[HTML]{F4B084} &  &  &  &  & \cellcolor[HTML]{F4B084} & \cellcolor[HTML]{BFBFBF} &  &  &  &  &  &  &  &  &  &  &  &  &  &  &  &  \\
\rowcolor[HTML]{F2F2F2} 
\cellcolor[HTML]{FCE4D6}C 7SVE V & \cellcolor[HTML]{8EA9DB} & \cellcolor[HTML]{8EA9DB} & \cellcolor[HTML]{8EA9DB} & \cellcolor[HTML]{8EA9DB} & \cellcolor[HTML]{8EA9DB} & \cellcolor[HTML]{8EA9DB} & \cellcolor[HTML]{8EA9DB} & \cellcolor[HTML]{8EA9DB} &  &  & \cellcolor[HTML]{8EA9DB} & \cellcolor[HTML]{8EA9DB} &  & \cellcolor[HTML]{8EA9DB} & \cellcolor[HTML]{8EA9DB} & \cellcolor[HTML]{8EA9DB} & \cellcolor[HTML]{BFBFBF} &  &  &  &  &  &  &  &  &  &  &  &  &  &  &  &  &  \\
\rowcolor[HTML]{F2F2F2} 
\cellcolor[HTML]{FCE4D6}SE 6SVE V &  &  &  &  &  & \cellcolor[HTML]{F4B084} & \cellcolor[HTML]{F4B084} & \cellcolor[HTML]{F4B084} &  &  &  &  &  &  &  & \cellcolor[HTML]{BFBFBF} &  &  &  &  &  &  &  &  &  &  &  &  &  &  &  &  &  &  \\
\rowcolor[HTML]{F2F2F2} 
\cellcolor[HTML]{FCE4D6}S 6SVE V &  &  &  &  &  & \cellcolor[HTML]{F4B084} & \cellcolor[HTML]{F4B084} &  &  &  &  &  &  &  & \cellcolor[HTML]{BFBFBF} &  &  &  &  &  &  &  &  &  &  &  &  &  &  &  &  &  &  &  \\
\rowcolor[HTML]{F2F2F2} 
\cellcolor[HTML]{FCE4D6}E 6SVE V &  &  &  &  &  & \cellcolor[HTML]{F4B084} &  &  &  &  &  &  &  & \cellcolor[HTML]{BFBFBF} &  &  &  &  &  &  &  &  &  &  &  &  &  &  &  &  &  &  &  &  \\
\rowcolor[HTML]{F2F2F2} 
\cellcolor[HTML]{FCE4D6}C 6SVE V &  & \cellcolor[HTML]{8EA9DB} & \cellcolor[HTML]{8EA9DB} & \cellcolor[HTML]{8EA9DB} &  & \cellcolor[HTML]{F4B084} & \cellcolor[HTML]{F4B084} & \cellcolor[HTML]{F4B084} &  &  & \cellcolor[HTML]{8EA9DB} & \cellcolor[HTML]{F4B084} & \cellcolor[HTML]{BFBFBF} &  &  &  &  &  &  &  &  &  &  &  &  &  &  &  &  &  &  &  &  &  \\
\rowcolor[HTML]{F2F2F2} 
\cellcolor[HTML]{FCE4D6}SE 5SVE V & \cellcolor[HTML]{8EA9DB} &  &  &  & \cellcolor[HTML]{8EA9DB} & \cellcolor[HTML]{F4B084} &  &  & \cellcolor[HTML]{8EA9DB} & \cellcolor[HTML]{8EA9DB} &  & \cellcolor[HTML]{BFBFBF} &  &  &  &  &  &  &  &  &  &  &  &  &  &  &  &  &  &  &  &  &  &  \\
\rowcolor[HTML]{F2F2F2} 
\cellcolor[HTML]{FCE4D6}S 5SVE V & \cellcolor[HTML]{F4B084} &  &  &  &  & \cellcolor[HTML]{F4B084} & \cellcolor[HTML]{F4B084} &  &  & \cellcolor[HTML]{8EA9DB} & \cellcolor[HTML]{BFBFBF} &  &  &  &  &  &  &  &  &  &  &  &  &  &  &  &  &  &  &  &  &  &  &  \\
\rowcolor[HTML]{F2F2F2} 
\cellcolor[HTML]{FCE4D6}E 5SVE V &  & \cellcolor[HTML]{8EA9DB} & \cellcolor[HTML]{8EA9DB} & \cellcolor[HTML]{8EA9DB} &  & \cellcolor[HTML]{F4B084} & \cellcolor[HTML]{F4B084} & \cellcolor[HTML]{F4B084} &  & \cellcolor[HTML]{BFBFBF} &  &  &  &  &  &  &  &  &  &  &  &  &  &  &  &  &  &  &  &  &  &  &  &  \\
\rowcolor[HTML]{F2F2F2} 
\cellcolor[HTML]{FCE4D6}C 5SVE V &  & \cellcolor[HTML]{8EA9DB} & \cellcolor[HTML]{F4B084} & \cellcolor[HTML]{F4B084} &  & \cellcolor[HTML]{F4B084} & \cellcolor[HTML]{F4B084} & \cellcolor[HTML]{F4B084} & \cellcolor[HTML]{BFBFBF} &  &  &  &  &  &  &  &  &  &  &  &  &  &  &  &  &  &  &  &  &  &  &  &  &  \\
\rowcolor[HTML]{F2F2F2} 
\cellcolor[HTML]{FCE4D6}SE 4SVE V & \cellcolor[HTML]{8EA9DB} &  &  &  & \cellcolor[HTML]{8EA9DB} &  &  & \cellcolor[HTML]{BFBFBF} &  &  &  &  &  &  &  &  &  &  &  &  &  &  &  &  &  &  &  &  &  &  &  &  &  &  \\
\rowcolor[HTML]{F2F2F2} 
\cellcolor[HTML]{FCE4D6}S 4SVE V & \cellcolor[HTML]{8EA9DB} &  &  &  & \cellcolor[HTML]{8EA9DB} &  & \cellcolor[HTML]{BFBFBF} &  &  &  &  &  &  &  &  &  &  &  &  &  &  &  &  &  &  &  &  &  &  &  &  &  &  &  \\
\rowcolor[HTML]{F2F2F2} 
\cellcolor[HTML]{FCE4D6}E 4SVE V & \cellcolor[HTML]{8EA9DB} &  &  &  & \cellcolor[HTML]{8EA9DB} & \cellcolor[HTML]{BFBFBF} &  &  &  &  &  &  &  &  &  &  &  &  &  &  &  &  &  &  &  &  &  &  &  &  &  &  &  &  \\
\rowcolor[HTML]{F2F2F2} 
\cellcolor[HTML]{FCE4D6}C 4SVE V &  & \cellcolor[HTML]{8EA9DB} & \cellcolor[HTML]{8EA9DB} & \cellcolor[HTML]{8EA9DB} & \cellcolor[HTML]{BFBFBF} &  &  &  &  &  &  &  &  &  &  &  &  &  &  &  &  &  &  &  &  &  &  &  &  &  &  &  &  &  \\
\rowcolor[HTML]{F2F2F2} 
\cellcolor[HTML]{FCE4D6}SE 3SVE V & \cellcolor[HTML]{F4B084} &  &  & \cellcolor[HTML]{BFBFBF} &  &  &  &  &  &  &  &  &  &  &  &  &  &  &  &  &  &  &  &  &  &  &  &  &  &  &  &  &  &  \\
\rowcolor[HTML]{F2F2F2} 
\cellcolor[HTML]{FCE4D6}S 3SVE V & \cellcolor[HTML]{F4B084} &  & \cellcolor[HTML]{BFBFBF} &  &  &  &  &  &  &  &  &  &  &  &  &  &  &  &  &  &  &  &  &  &  &  &  &  &  &  &  &  &  &  \\
\rowcolor[HTML]{F2F2F2} 
\cellcolor[HTML]{FCE4D6}E 3SVE V & \cellcolor[HTML]{F4B084} & \cellcolor[HTML]{BFBFBF} &  &  &  &  &  &  &  &  &  &  &  &  &  &  &  &  &  &  &  &  &  &  &  &  &  &  &  &  &  &  &  &  \\ \hline
\end{tabular}%
}
\end{table}

\begin{table}[ht]
\caption{Ball Retention Mann-Whitney Row/Col comparison}
\label{tab:ball_grip}
\resizebox{\columnwidth}{!}{%
\begin{tabular}{|c|llllllllllllllllllllllllllllllllll|}
\hline
\rowcolor[HTML]{D9E1F2}
\cellcolor[HTML]{FFFFFF} & \rotatebox[origin=c]{90}{C 3SVE V} & \rotatebox[origin=c]{90}{E 3SVE V} & \rotatebox[origin=c]{90}{S 3SVE V} & \rotatebox[origin=c]{90}{SE 3SVE V} & \rotatebox[origin=c]{90}{C 4SVE V} & \rotatebox[origin=c]{90}{E 4SVE V} & \rotatebox[origin=c]{90}{S 4SVE V} & \rotatebox[origin=c]{90}{SE 4SVE V} & \rotatebox[origin=c]{90}{C 5SVE V} & \rotatebox[origin=c]{90}{E 5SVE V} & \rotatebox[origin=c]{90}{S 5SVE V} & \rotatebox[origin=c]{90}{SE 5SVE V} & \rotatebox[origin=c]{90}{C 6SVE V} & \rotatebox[origin=c]{90}{E 6SVE V} & \rotatebox[origin=c]{90}{S 6SVE V} & \rotatebox[origin=c]{90}{SE 6SVE V} & \rotatebox[origin=c]{90}{C 7SVE V} & \rotatebox[origin=c]{90}{E 7SVE V} & \rotatebox[origin=c]{90}{S 7SVE V} & \rotatebox[origin=c]{90}{SE 7SVE V} & \rotatebox[origin=c]{90}{S 3 G} & \rotatebox[origin=c]{90}{S 6 G} & \rotatebox[origin=c]{90}{S 8 G} & \rotatebox[origin=c]{90}{S 3 SP} & \rotatebox[origin=c]{90}{C 3 HP} & \rotatebox[origin=c]{90}{C 7 HP} & \rotatebox[origin=c]{90}{S 7 HP} & \rotatebox[origin=c]{90}{Rice} & \rotatebox[origin=c]{90}{Coffee} & \rotatebox[origin=c]{90}{Polystyrene} & \rotatebox[origin=c]{90}{Rubber} & \rotatebox[origin=c]{90}{C 4 AG70} & \rotatebox[origin=c]{90}{C 4 AG30} & \rotatebox[origin=c]{90}{SEBS30} \\ \hline
\rowcolor[HTML]{8EA9DB} 
\cellcolor[HTML]{FCE4D6}SEBS0 & \cellcolor[HTML]{F2F2F2} &  & \cellcolor[HTML]{F2F2F2} &  &  &  &  & \cellcolor[HTML]{F4B084} & \cellcolor[HTML]{F2F2F2} & \cellcolor[HTML]{F2F2F2} & \cellcolor[HTML]{F2F2F2} &  & \cellcolor[HTML]{F2F2F2} &  & \cellcolor[HTML]{F2F2F2} &  &  &  &  &  & \cellcolor[HTML]{F4B084} & \cellcolor[HTML]{F4B084} &  &  &  & \cellcolor[HTML]{F2F2F2} &  &  & \cellcolor[HTML]{F4B084} & \cellcolor[HTML]{F4B084} & \cellcolor[HTML]{F4B084} & \cellcolor[HTML]{F4B084} & \cellcolor[HTML]{F4B084} &  \\
\cellcolor[HTML]{FCE4D6}SEBS30 & \cellcolor[HTML]{F4B084} & \cellcolor[HTML]{F2F2F2} & \cellcolor[HTML]{F4B084} & \cellcolor[HTML]{F2F2F2} & \cellcolor[HTML]{F2F2F2} & \cellcolor[HTML]{F4B084} & \cellcolor[HTML]{F2F2F2} & \cellcolor[HTML]{F4B084} & \cellcolor[HTML]{F2F2F2} & \cellcolor[HTML]{8EA9DB} & \cellcolor[HTML]{F2F2F2} & \cellcolor[HTML]{F2F2F2} & \cellcolor[HTML]{8EA9DB} & \cellcolor[HTML]{F2F2F2} & \cellcolor[HTML]{F2F2F2} & \cellcolor[HTML]{F2F2F2} & \cellcolor[HTML]{F2F2F2} & \cellcolor[HTML]{F2F2F2} & \cellcolor[HTML]{F2F2F2} & \cellcolor[HTML]{F2F2F2} & \cellcolor[HTML]{F4B084} & \cellcolor[HTML]{F4B084} & \cellcolor[HTML]{F2F2F2} & \cellcolor[HTML]{F2F2F2} & \cellcolor[HTML]{F2F2F2} & \cellcolor[HTML]{F4B084} & \cellcolor[HTML]{8EA9DB} & \cellcolor[HTML]{F4B084} & \cellcolor[HTML]{F4B084} & \cellcolor[HTML]{F4B084} & \cellcolor[HTML]{F4B084} & \cellcolor[HTML]{F4B084} & \cellcolor[HTML]{F4B084} & \cellcolor[HTML]{BFBFBF} \\
\rowcolor[HTML]{8EA9DB} 
\cellcolor[HTML]{FCE4D6}C 4 AG30 &  &  &  &  &  &  &  &  &  &  &  &  &  &  &  &  &  &  &  &  & \cellcolor[HTML]{F4B084} &  &  &  &  &  &  &  &  & \cellcolor[HTML]{F2F2F2} & \cellcolor[HTML]{F2F2F2} &  & \cellcolor[HTML]{BFBFBF} & \cellcolor[HTML]{F2F2F2} \\
\rowcolor[HTML]{8EA9DB} 
\cellcolor[HTML]{FCE4D6}C 4 AG70 & \cellcolor[HTML]{F2F2F2} &  &  &  &  &  &  &  &  &  &  &  &  &  &  &  &  &  &  &  & \cellcolor[HTML]{F4B084} &  &  &  &  &  &  &  & \cellcolor[HTML]{F2F2F2} &  &  & \cellcolor[HTML]{BFBFBF} & \cellcolor[HTML]{F2F2F2} & \cellcolor[HTML]{F2F2F2} \\
\rowcolor[HTML]{8EA9DB} 
\cellcolor[HTML]{FCE4D6}Rubber &  &  &  &  &  &  &  &  &  &  &  &  &  &  &  &  &  &  &  &  & \cellcolor[HTML]{F4B084} &  &  &  &  &  &  &  &  & \cellcolor[HTML]{F2F2F2} & \cellcolor[HTML]{BFBFBF} & \cellcolor[HTML]{F2F2F2} & \cellcolor[HTML]{F2F2F2} & \cellcolor[HTML]{F2F2F2} \\
\rowcolor[HTML]{8EA9DB} 
\cellcolor[HTML]{FCE4D6}Polystyrene &  &  &  &  &  &  &  &  &  &  &  &  &  &  &  &  &  &  &  &  & \cellcolor[HTML]{F4B084} &  &  &  &  &  &  &  &  & \cellcolor[HTML]{BFBFBF} & \cellcolor[HTML]{F2F2F2} & \cellcolor[HTML]{F2F2F2} & \cellcolor[HTML]{F2F2F2} & \cellcolor[HTML]{F2F2F2} \\
\rowcolor[HTML]{8EA9DB} 
\cellcolor[HTML]{FCE4D6}Coffee & \cellcolor[HTML]{F2F2F2} &  & \cellcolor[HTML]{F2F2F2} &  &  &  &  & \cellcolor[HTML]{F4B084} &  &  &  &  & \cellcolor[HTML]{F2F2F2} &  &  &  &  &  &  &  & \cellcolor[HTML]{F4B084} & \cellcolor[HTML]{F4B084} &  &  &  & \cellcolor[HTML]{F2F2F2} &  &  & \cellcolor[HTML]{BFBFBF} & \cellcolor[HTML]{F2F2F2} & \cellcolor[HTML]{F2F2F2} & \cellcolor[HTML]{F2F2F2} & \cellcolor[HTML]{F2F2F2} & \cellcolor[HTML]{F2F2F2} \\
\rowcolor[HTML]{F2F2F2} 
\cellcolor[HTML]{FCE4D6}Rice & \cellcolor[HTML]{F4B084} & \cellcolor[HTML]{8EA9DB} & \cellcolor[HTML]{8EA9DB} &  & \cellcolor[HTML]{8EA9DB} &  &  & \cellcolor[HTML]{F4B084} & \cellcolor[HTML]{8EA9DB} & \cellcolor[HTML]{8EA9DB} & \cellcolor[HTML]{8EA9DB} &  & \cellcolor[HTML]{8EA9DB} & \cellcolor[HTML]{8EA9DB} &  &  & \cellcolor[HTML]{8EA9DB} & \cellcolor[HTML]{8EA9DB} & \cellcolor[HTML]{8EA9DB} &  & \cellcolor[HTML]{F4B084} &  &  &  &  & \cellcolor[HTML]{F4B084} &  & \cellcolor[HTML]{BFBFBF} &  &  &  &  &  &  \\
\rowcolor[HTML]{F2F2F2} 
\cellcolor[HTML]{FCE4D6}S 7 HP & \cellcolor[HTML]{F4B084} & \cellcolor[HTML]{8EA9DB} & \cellcolor[HTML]{F4B084} &  &  &  &  & \cellcolor[HTML]{F4B084} &  & \cellcolor[HTML]{8EA9DB} & \cellcolor[HTML]{8EA9DB} &  & \cellcolor[HTML]{8EA9DB} &  &  &  &  & \cellcolor[HTML]{8EA9DB} &  &  & \cellcolor[HTML]{F4B084} &  &  &  &  & \cellcolor[HTML]{F4B084} & \cellcolor[HTML]{BFBFBF} &  &  &  &  &  &  &  \\
\cellcolor[HTML]{FCE4D6}C 7 HP & \cellcolor[HTML]{F2F2F2} & \cellcolor[HTML]{8EA9DB} & \cellcolor[HTML]{F2F2F2} & \cellcolor[HTML]{8EA9DB} & \cellcolor[HTML]{8EA9DB} & \cellcolor[HTML]{8EA9DB} & \cellcolor[HTML]{8EA9DB} & \cellcolor[HTML]{F4B084} & \cellcolor[HTML]{F2F2F2} & \cellcolor[HTML]{F2F2F2} & \cellcolor[HTML]{F2F2F2} & \cellcolor[HTML]{8EA9DB} & \cellcolor[HTML]{F2F2F2} & \cellcolor[HTML]{8EA9DB} & \cellcolor[HTML]{8EA9DB} & \cellcolor[HTML]{8EA9DB} & \cellcolor[HTML]{8EA9DB} & \cellcolor[HTML]{8EA9DB} & \cellcolor[HTML]{8EA9DB} & \cellcolor[HTML]{8EA9DB} & \cellcolor[HTML]{F4B084} & \cellcolor[HTML]{F4B084} & \cellcolor[HTML]{8EA9DB} & \cellcolor[HTML]{8EA9DB} & \cellcolor[HTML]{8EA9DB} & \cellcolor[HTML]{BFBFBF} & \cellcolor[HTML]{F2F2F2} & \cellcolor[HTML]{F2F2F2} & \cellcolor[HTML]{F2F2F2} & \cellcolor[HTML]{F2F2F2} & \cellcolor[HTML]{F2F2F2} & \cellcolor[HTML]{F2F2F2} & \cellcolor[HTML]{F2F2F2} & \cellcolor[HTML]{F2F2F2} \\
\rowcolor[HTML]{F2F2F2} 
\cellcolor[HTML]{FCE4D6}C 3 HP & \cellcolor[HTML]{F4B084} &  & \cellcolor[HTML]{F4B084} &  &  &  &  & \cellcolor[HTML]{F4B084} &  &  &  &  & \cellcolor[HTML]{F4B084} &  &  &  &  &  &  &  & \cellcolor[HTML]{F4B084} & \cellcolor[HTML]{F4B084} &  &  & \cellcolor[HTML]{BFBFBF} &  &  &  &  &  &  &  &  &  \\
\rowcolor[HTML]{F2F2F2} 
\cellcolor[HTML]{FCE4D6}S 3 SP & \cellcolor[HTML]{F4B084} &  & \cellcolor[HTML]{F4B084} &  &  &  &  & \cellcolor[HTML]{F4B084} &  & \cellcolor[HTML]{F4B084} &  &  & \cellcolor[HTML]{F4B084} &  &  &  &  &  &  &  & \cellcolor[HTML]{F4B084} & \cellcolor[HTML]{F4B084} &  & \cellcolor[HTML]{BFBFBF} &  &  &  &  &  &  &  &  &  &  \\
\rowcolor[HTML]{F2F2F2} 
\cellcolor[HTML]{FCE4D6}S 8 G & \cellcolor[HTML]{F4B084} &  & \cellcolor[HTML]{F4B084} &  &  &  &  & \cellcolor[HTML]{F4B084} &  & \cellcolor[HTML]{F4B084} &  &  & \cellcolor[HTML]{F4B084} &  &  &  &  &  &  &  & \cellcolor[HTML]{F4B084} & \cellcolor[HTML]{F4B084} & \cellcolor[HTML]{BFBFBF} &  &  &  &  &  &  &  &  &  &  &  \\
\rowcolor[HTML]{8EA9DB} 
\cellcolor[HTML]{FCE4D6}S 6 G &  &  &  &  &  & \cellcolor[HTML]{F2F2F2} &  & \cellcolor[HTML]{F2F2F2} &  &  &  &  &  &  &  &  &  &  &  &  & \cellcolor[HTML]{F4B084} & \cellcolor[HTML]{BFBFBF} & \cellcolor[HTML]{F2F2F2} & \cellcolor[HTML]{F2F2F2} & \cellcolor[HTML]{F2F2F2} & \cellcolor[HTML]{F2F2F2} & \cellcolor[HTML]{F2F2F2} & \cellcolor[HTML]{F2F2F2} & \cellcolor[HTML]{F2F2F2} & \cellcolor[HTML]{F2F2F2} & \cellcolor[HTML]{F2F2F2} & \cellcolor[HTML]{F2F2F2} & \cellcolor[HTML]{F2F2F2} & \cellcolor[HTML]{F2F2F2} \\
\rowcolor[HTML]{8EA9DB} 
\cellcolor[HTML]{FCE4D6}S 3 G &  &  &  &  &  &  &  & \cellcolor[HTML]{F2F2F2} &  &  &  &  &  &  &  &  &  &  &  &  & \cellcolor[HTML]{BFBFBF} & \cellcolor[HTML]{F2F2F2} & \cellcolor[HTML]{F2F2F2} & \cellcolor[HTML]{F2F2F2} & \cellcolor[HTML]{F2F2F2} & \cellcolor[HTML]{F2F2F2} & \cellcolor[HTML]{F2F2F2} & \cellcolor[HTML]{F2F2F2} & \cellcolor[HTML]{F2F2F2} & \cellcolor[HTML]{F2F2F2} & \cellcolor[HTML]{F2F2F2} & \cellcolor[HTML]{F2F2F2} & \cellcolor[HTML]{F2F2F2} & \cellcolor[HTML]{F2F2F2} \\
\rowcolor[HTML]{F2F2F2} 
\cellcolor[HTML]{FCE4D6}SE 7SVE V & \cellcolor[HTML]{F4B084} &  &  &  &  &  &  & \cellcolor[HTML]{F4B084} &  &  &  &  &  &  &  &  &  &  &  & \cellcolor[HTML]{BFBFBF} &  &  &  &  &  &  &  &  &  &  &  &  &  &  \\
\rowcolor[HTML]{F2F2F2} 
\cellcolor[HTML]{FCE4D6}S 7SVE V & \cellcolor[HTML]{F4B084} &  &  &  &  & \cellcolor[HTML]{F4B084} &  & \cellcolor[HTML]{F4B084} &  &  &  &  &  &  &  &  &  &  & \cellcolor[HTML]{BFBFBF} &  &  &  &  &  &  &  &  &  &  &  &  &  &  &  \\
\rowcolor[HTML]{F2F2F2} 
\cellcolor[HTML]{FCE4D6}E 7SVE V & \cellcolor[HTML]{F4B084} &  &  &  &  & \cellcolor[HTML]{F4B084} &  & \cellcolor[HTML]{F4B084} &  &  &  &  &  &  &  &  &  & \cellcolor[HTML]{BFBFBF} &  &  &  &  &  &  &  &  &  &  &  &  &  &  &  &  \\
\rowcolor[HTML]{F2F2F2} 
\cellcolor[HTML]{FCE4D6}C 7SVE V & \cellcolor[HTML]{F4B084} &  & \cellcolor[HTML]{F4B084} &  &  &  &  & \cellcolor[HTML]{F4B084} &  & \cellcolor[HTML]{F4B084} &  &  & \cellcolor[HTML]{F4B084} &  &  &  & \cellcolor[HTML]{BFBFBF} &  &  &  &  &  &  &  &  &  &  &  &  &  &  &  &  &  \\
\rowcolor[HTML]{F2F2F2} 
\cellcolor[HTML]{FCE4D6}SE 6SVE V & \cellcolor[HTML]{F4B084} &  & \cellcolor[HTML]{F4B084} &  &  &  &  & \cellcolor[HTML]{F4B084} &  &  &  &  &  &  &  & \cellcolor[HTML]{BFBFBF} &  &  &  &  &  &  &  &  &  &  &  &  &  &  &  &  &  &  \\
\rowcolor[HTML]{F2F2F2} 
\cellcolor[HTML]{FCE4D6}S 6SVE V & \cellcolor[HTML]{F4B084} &  &  &  &  &  &  & \cellcolor[HTML]{F4B084} &  &  &  &  &  &  & \cellcolor[HTML]{BFBFBF} &  &  &  &  &  &  &  &  &  &  &  &  &  &  &  &  &  &  &  \\
\rowcolor[HTML]{F2F2F2} 
\cellcolor[HTML]{FCE4D6}E 6SVE V & \cellcolor[HTML]{F4B084} &  & \cellcolor[HTML]{F4B084} &  &  & \cellcolor[HTML]{F4B084} &  & \cellcolor[HTML]{F4B084} &  &  &  &  &  & \cellcolor[HTML]{BFBFBF} &  &  &  &  &  &  &  &  &  &  &  &  &  &  &  &  &  &  &  &  \\
\rowcolor[HTML]{F2F2F2} 
\cellcolor[HTML]{FCE4D6}C 6SVE V &  &  &  & \cellcolor[HTML]{8EA9DB} &  & \cellcolor[HTML]{F4B084} & \cellcolor[HTML]{8EA9DB} & \cellcolor[HTML]{F4B084} &  &  &  &  & \cellcolor[HTML]{BFBFBF} &  &  &  &  &  &  &  &  &  &  &  &  &  &  &  &  &  &  &  &  &  \\
\rowcolor[HTML]{F2F2F2} 
\cellcolor[HTML]{FCE4D6}SE 5SVE V & \cellcolor[HTML]{F4B084} &  &  &  &  &  &  & \cellcolor[HTML]{F4B084} &  &  &  & \cellcolor[HTML]{BFBFBF} &  &  &  &  &  &  &  &  &  &  &  &  &  &  &  &  &  &  &  &  &  &  \\
\rowcolor[HTML]{F2F2F2} 
\cellcolor[HTML]{FCE4D6}S 5SVE V & \cellcolor[HTML]{F4B084} &  &  &  &  & \cellcolor[HTML]{F4B084} &  & \cellcolor[HTML]{F4B084} &  &  & \cellcolor[HTML]{BFBFBF} &  &  &  &  &  &  &  &  &  &  &  &  &  &  &  &  &  &  &  &  &  &  &  \\
\rowcolor[HTML]{F2F2F2} 
\cellcolor[HTML]{FCE4D6}E 5SVE V & \cellcolor[HTML]{F4B084} &  &  &  &  & \cellcolor[HTML]{F4B084} & \cellcolor[HTML]{8EA9DB} & \cellcolor[HTML]{F4B084} &  & \cellcolor[HTML]{BFBFBF} &  &  &  &  &  &  &  &  &  &  &  &  &  &  &  &  &  &  &  &  &  &  &  &  \\
\rowcolor[HTML]{F2F2F2} 
\cellcolor[HTML]{FCE4D6}C 5SVE V & \cellcolor[HTML]{F4B084} &  &  &  &  & \cellcolor[HTML]{F4B084} &  & \cellcolor[HTML]{F4B084} & \cellcolor[HTML]{BFBFBF} &  &  &  &  &  &  &  &  &  &  &  &  &  &  &  &  &  &  &  &  &  &  &  &  &  \\
\rowcolor[HTML]{F2F2F2} 
\cellcolor[HTML]{FCE4D6}SE 4SVE V & \cellcolor[HTML]{8EA9DB} & \cellcolor[HTML]{8EA9DB} & \cellcolor[HTML]{8EA9DB} & \cellcolor[HTML]{8EA9DB} & \cellcolor[HTML]{8EA9DB} & \cellcolor[HTML]{8EA9DB} & \cellcolor[HTML]{8EA9DB} & \cellcolor[HTML]{BFBFBF} &  &  &  &  &  &  &  &  &  &  &  &  &  &  &  &  &  &  &  &  &  &  &  &  &  &  \\
\rowcolor[HTML]{F2F2F2} 
\cellcolor[HTML]{FCE4D6}S 4SVE V & \cellcolor[HTML]{F4B084} &  & \cellcolor[HTML]{F4B084} &  &  &  & \cellcolor[HTML]{BFBFBF} &  &  &  &  &  &  &  &  &  &  &  &  &  &  &  &  &  &  &  &  &  &  &  &  &  &  &  \\
\rowcolor[HTML]{F2F2F2} 
\cellcolor[HTML]{FCE4D6}E 4SVE V & \cellcolor[HTML]{F4B084} & \cellcolor[HTML]{8EA9DB} & \cellcolor[HTML]{8EA9DB} &  &  & \cellcolor[HTML]{BFBFBF} &  &  &  &  &  &  &  &  &  &  &  &  &  &  &  &  &  &  &  &  &  &  &  &  &  &  &  &  \\
\rowcolor[HTML]{F2F2F2} 
\cellcolor[HTML]{FCE4D6}C 4SVE V & \cellcolor[HTML]{F4B084} &  &  &  & \cellcolor[HTML]{BFBFBF} &  &  &  &  &  &  &  &  &  &  &  &  &  &  &  &  &  &  &  &  &  &  &  &  &  &  &  &  &  \\
\rowcolor[HTML]{F2F2F2} 
\cellcolor[HTML]{FCE4D6}SE 3SVE V & \cellcolor[HTML]{F4B084} &  &  & \cellcolor[HTML]{BFBFBF} &  &  &  &  &  &  &  &  &  &  &  &  &  &  &  &  &  &  &  &  &  &  &  &  &  &  &  &  &  &  \\
\rowcolor[HTML]{F2F2F2} 
\cellcolor[HTML]{FCE4D6}S 3SVE V & \cellcolor[HTML]{F4B084} & \cellcolor[HTML]{8EA9DB} & \cellcolor[HTML]{BFBFBF} &  &  &  &  &  &  &  &  &  &  &  &  &  &  &  &  &  &  &  &  &  &  &  &  &  &  &  &  &  &  &  \\
\rowcolor[HTML]{F2F2F2} 
\cellcolor[HTML]{FCE4D6}E 3SVE V & \cellcolor[HTML]{F4B084} & \cellcolor[HTML]{BFBFBF} &  &  &  &  &  &  &  &  &  &  &  &  &  &  &  &  &  &  &  &  &  &  &  &  &  &  &  &  &  &  &  &  \\ \hline
\end{tabular}%
}
\end{table}

\begin{table}[ht]
\caption{Coin Retention Mann-Whitney Row/Col comparison}
\label{tab:coin_grip}
\resizebox{\columnwidth}{!}{%
    \begin{tabular}{|c|llllllllllllllllllllllllllllllllll|}
\hline
\rowcolor[HTML]{D9E1F2}
\cellcolor[HTML]{FFFFFF} & \rotatebox[origin=c]{90}{C 3SVE V} & \rotatebox[origin=c]{90}{E 3SVE V} & \rotatebox[origin=c]{90}{S 3SVE V} & \rotatebox[origin=c]{90}{SE 3SVE V} & \rotatebox[origin=c]{90}{C 4SVE V} & \rotatebox[origin=c]{90}{E 4SVE V} & \rotatebox[origin=c]{90}{S 4SVE V} & \rotatebox[origin=c]{90}{SE 4SVE V} & \rotatebox[origin=c]{90}{C 5SVE V} & \rotatebox[origin=c]{90}{E 5SVE V} & \rotatebox[origin=c]{90}{S 5SVE V} & \rotatebox[origin=c]{90}{SE 5SVE V} & \rotatebox[origin=c]{90}{C 6SVE V} & \rotatebox[origin=c]{90}{E 6SVE V} & \rotatebox[origin=c]{90}{S 6SVE V} & \rotatebox[origin=c]{90}{SE 6SVE V} & \rotatebox[origin=c]{90}{C 7SVE V} & \rotatebox[origin=c]{90}{E 7SVE V} & \rotatebox[origin=c]{90}{S 7SVE V} & \rotatebox[origin=c]{90}{SE 7SVE V} & \rotatebox[origin=c]{90}{S 3 G} & \rotatebox[origin=c]{90}{S 6 G} & \rotatebox[origin=c]{90}{S 8 G} & \rotatebox[origin=c]{90}{S 3 SP} & \rotatebox[origin=c]{90}{C 3 HP} & \rotatebox[origin=c]{90}{C 7 HP} & \rotatebox[origin=c]{90}{S 7 HP} & \rotatebox[origin=c]{90}{Rice} & \rotatebox[origin=c]{90}{Coffee} & \rotatebox[origin=c]{90}{Polystyrene} & \rotatebox[origin=c]{90}{Rubber} & \rotatebox[origin=c]{90}{C 4 AG70} & \rotatebox[origin=c]{90}{C 4 AG30} & \rotatebox[origin=c]{90}{SEBS30} \\ \hline
\rowcolor[HTML]{F2F2F2} 
\cellcolor[HTML]{FCE4D6}SEBS0 & \cellcolor[HTML]{F4B084} &  &  &  & \cellcolor[HTML]{8EA9DB} &  &  &  & \cellcolor[HTML]{8EA9DB} & \cellcolor[HTML]{8EA9DB} &  &  & \cellcolor[HTML]{8EA9DB} &  &  &  & \cellcolor[HTML]{F4B084} & \cellcolor[HTML]{8EA9DB} & \cellcolor[HTML]{8EA9DB} & \cellcolor[HTML]{F4B084} &  &  & \cellcolor[HTML]{8EA9DB} &  & \cellcolor[HTML]{F4B084} &  &  &  &  & \cellcolor[HTML]{F4B084} & \cellcolor[HTML]{F4B084} & \cellcolor[HTML]{F4B084} & \cellcolor[HTML]{F4B084} & \cellcolor[HTML]{F4B084} \\
\rowcolor[HTML]{8EA9DB} 
\cellcolor[HTML]{FCE4D6}SEBS30 &  & \cellcolor[HTML]{F2F2F2} &  & \cellcolor[HTML]{F2F2F2} &  & \cellcolor[HTML]{F2F2F2} & \cellcolor[HTML]{F2F2F2} & \cellcolor[HTML]{F2F2F2} &  &  &  &  &  & \cellcolor[HTML]{F2F2F2} &  &  & \cellcolor[HTML]{F4B084} &  &  & \cellcolor[HTML]{F4B084} & \cellcolor[HTML]{F2F2F2} &  &  &  & \cellcolor[HTML]{F4B084} & \cellcolor[HTML]{F2F2F2} & \cellcolor[HTML]{F2F2F2} &  & \cellcolor[HTML]{F2F2F2} &  & \cellcolor[HTML]{F4B084} & \cellcolor[HTML]{F4B084} &  & \cellcolor[HTML]{BFBFBF} \\
\rowcolor[HTML]{F2F2F2} 
\cellcolor[HTML]{FCE4D6}C 4 AG30 &  & \cellcolor[HTML]{8EA9DB} & \cellcolor[HTML]{8EA9DB} & \cellcolor[HTML]{8EA9DB} &  & \cellcolor[HTML]{F4B084} & \cellcolor[HTML]{F4B084} & \cellcolor[HTML]{F4B084} &  &  & \cellcolor[HTML]{8EA9DB} & \cellcolor[HTML]{F4B084} &  &  &  &  & \cellcolor[HTML]{F4B084} &  &  & \cellcolor[HTML]{F4B084} & \cellcolor[HTML]{F4B084} &  &  &  & \cellcolor[HTML]{F4B084} &  &  &  &  &  & \cellcolor[HTML]{F4B084} & \cellcolor[HTML]{F4B084} & \cellcolor[HTML]{BFBFBF} &  \\
\rowcolor[HTML]{8EA9DB} 
\cellcolor[HTML]{FCE4D6}C 4 AG70 &  &  &  &  &  &  &  &  &  &  &  &  &  &  &  &  &  &  &  &  &  &  &  &  & \cellcolor[HTML]{F2F2F2} &  &  &  &  &  & \cellcolor[HTML]{F2F2F2} & \cellcolor[HTML]{BFBFBF} & \cellcolor[HTML]{F2F2F2} & \cellcolor[HTML]{F2F2F2} \\
\rowcolor[HTML]{8EA9DB} 
\cellcolor[HTML]{FCE4D6}Rubber &  &  &  &  &  &  &  &  &  &  &  &  &  &  &  &  &  &  &  &  &  &  &  &  & \cellcolor[HTML]{F2F2F2} &  &  &  &  &  & \cellcolor[HTML]{BFBFBF} & \cellcolor[HTML]{F2F2F2} & \cellcolor[HTML]{F2F2F2} & \cellcolor[HTML]{F2F2F2} \\
\cellcolor[HTML]{FCE4D6}Polystyrene & \cellcolor[HTML]{F2F2F2} & \cellcolor[HTML]{8EA9DB} & \cellcolor[HTML]{8EA9DB} & \cellcolor[HTML]{8EA9DB} & \cellcolor[HTML]{F2F2F2} & \cellcolor[HTML]{F4B084} & \cellcolor[HTML]{8EA9DB} & \cellcolor[HTML]{8EA9DB} & \cellcolor[HTML]{F2F2F2} & \cellcolor[HTML]{F2F2F2} & \cellcolor[HTML]{8EA9DB} & \cellcolor[HTML]{8EA9DB} & \cellcolor[HTML]{F2F2F2} & \cellcolor[HTML]{8EA9DB} & \cellcolor[HTML]{8EA9DB} & \cellcolor[HTML]{F2F2F2} & \cellcolor[HTML]{F2F2F2} & \cellcolor[HTML]{F2F2F2} & \cellcolor[HTML]{F2F2F2} & \cellcolor[HTML]{F4B084} & \cellcolor[HTML]{8EA9DB} & \cellcolor[HTML]{8EA9DB} & \cellcolor[HTML]{8EA9DB} & \cellcolor[HTML]{F2F2F2} & \cellcolor[HTML]{F4B084} & \cellcolor[HTML]{F2F2F2} & \cellcolor[HTML]{8EA9DB} & \cellcolor[HTML]{F2F2F2} & \cellcolor[HTML]{8EA9DB} & \cellcolor[HTML]{BFBFBF} & \cellcolor[HTML]{F2F2F2} & \cellcolor[HTML]{F2F2F2} & \cellcolor[HTML]{F2F2F2} & \cellcolor[HTML]{F2F2F2} \\
\rowcolor[HTML]{F2F2F2} 
\cellcolor[HTML]{FCE4D6}Coffee &  &  &  &  &  & \cellcolor[HTML]{F4B084} &  &  &  &  &  &  &  &  &  &  & \cellcolor[HTML]{F4B084} &  &  & \cellcolor[HTML]{F4B084} &  &  &  &  & \cellcolor[HTML]{F4B084} &  &  &  & \cellcolor[HTML]{BFBFBF} &  &  &  &  &  \\
\rowcolor[HTML]{F2F2F2} 
\cellcolor[HTML]{FCE4D6}Rice &  &  &  &  &  & \cellcolor[HTML]{F4B084} & \cellcolor[HTML]{F4B084} & \cellcolor[HTML]{F4B084} &  &  &  &  &  &  &  &  & \cellcolor[HTML]{F4B084} &  &  & \cellcolor[HTML]{F4B084} &  &  &  &  & \cellcolor[HTML]{F4B084} &  &  & \cellcolor[HTML]{BFBFBF} &  &  &  &  &  &  \\
\rowcolor[HTML]{F2F2F2} 
\cellcolor[HTML]{FCE4D6}S 7 HP &  &  &  &  &  &  &  &  &  &  &  &  &  &  &  &  & \cellcolor[HTML]{F4B084} &  &  & \cellcolor[HTML]{F4B084} &  &  &  &  & \cellcolor[HTML]{F4B084} &  & \cellcolor[HTML]{BFBFBF} &  &  &  &  &  &  &  \\
\rowcolor[HTML]{F2F2F2} 
\cellcolor[HTML]{FCE4D6}C 7 HP &  &  &  &  &  &  &  &  &  &  &  &  &  &  &  &  & \cellcolor[HTML]{F4B084} &  &  & \cellcolor[HTML]{F4B084} &  &  &  &  & \cellcolor[HTML]{F4B084} & \cellcolor[HTML]{BFBFBF} &  &  &  &  &  &  &  &  \\
\rowcolor[HTML]{8EA9DB} 
\cellcolor[HTML]{FCE4D6}C 3 HP &  &  &  &  &  &  &  &  & \cellcolor[HTML]{F2F2F2} & \cellcolor[HTML]{F2F2F2} &  &  & \cellcolor[HTML]{F2F2F2} &  &  &  & \cellcolor[HTML]{F2F2F2} &  & \cellcolor[HTML]{F2F2F2} & \cellcolor[HTML]{F2F2F2} &  &  &  &  & \cellcolor[HTML]{BFBFBF} & \cellcolor[HTML]{F2F2F2} & \cellcolor[HTML]{F2F2F2} & \cellcolor[HTML]{F2F2F2} & \cellcolor[HTML]{F2F2F2} & \cellcolor[HTML]{F2F2F2} & \cellcolor[HTML]{F2F2F2} & \cellcolor[HTML]{F2F2F2} & \cellcolor[HTML]{F2F2F2} & \cellcolor[HTML]{F2F2F2} \\
\rowcolor[HTML]{F2F2F2} 
\cellcolor[HTML]{FCE4D6}S 3 SP &  &  &  &  &  & \cellcolor[HTML]{F4B084} & \cellcolor[HTML]{F4B084} &  &  &  &  &  &  &  &  &  & \cellcolor[HTML]{F4B084} &  &  & \cellcolor[HTML]{F4B084} &  &  &  & \cellcolor[HTML]{BFBFBF} &  &  &  &  &  &  &  &  &  &  \\
\rowcolor[HTML]{F2F2F2} 
\cellcolor[HTML]{FCE4D6}S 8 G &  &  &  &  &  & \cellcolor[HTML]{F4B084} & \cellcolor[HTML]{F4B084} & \cellcolor[HTML]{F4B084} &  &  &  &  &  &  &  &  & \cellcolor[HTML]{F4B084} &  &  & \cellcolor[HTML]{F4B084} & \cellcolor[HTML]{F4B084} &  & \cellcolor[HTML]{BFBFBF} &  &  &  &  &  &  &  &  &  &  &  \\
\rowcolor[HTML]{F2F2F2} 
\cellcolor[HTML]{FCE4D6}S 6 G &  &  &  &  &  &  &  &  &  & \cellcolor[HTML]{F4B084} &  &  & \cellcolor[HTML]{F4B084} &  &  &  & \cellcolor[HTML]{F4B084} &  & \cellcolor[HTML]{F4B084} & \cellcolor[HTML]{F4B084} &  & \cellcolor[HTML]{BFBFBF} &  &  &  &  &  &  &  &  &  &  &  &  \\
\rowcolor[HTML]{F2F2F2} 
\cellcolor[HTML]{FCE4D6}S 3 G & \cellcolor[HTML]{F4B084} &  &  &  & \cellcolor[HTML]{8EA9DB} &  &  &  & \cellcolor[HTML]{8EA9DB} & \cellcolor[HTML]{8EA9DB} &  &  & \cellcolor[HTML]{8EA9DB} &  &  & \cellcolor[HTML]{8EA9DB} & \cellcolor[HTML]{F4B084} & \cellcolor[HTML]{8EA9DB} & \cellcolor[HTML]{8EA9DB} & \cellcolor[HTML]{F4B084} & \cellcolor[HTML]{BFBFBF} &  &  &  &  &  &  &  &  &  &  &  &  &  \\
\rowcolor[HTML]{F2F2F2} 
\cellcolor[HTML]{FCE4D6}SE 7SVE V & \cellcolor[HTML]{8EA9DB} & \cellcolor[HTML]{8EA9DB} & \cellcolor[HTML]{8EA9DB} & \cellcolor[HTML]{8EA9DB} & \cellcolor[HTML]{8EA9DB} & \cellcolor[HTML]{8EA9DB} & \cellcolor[HTML]{8EA9DB} & \cellcolor[HTML]{8EA9DB} & \cellcolor[HTML]{8EA9DB} &  & \cellcolor[HTML]{8EA9DB} & \cellcolor[HTML]{8EA9DB} &  & \cellcolor[HTML]{8EA9DB} & \cellcolor[HTML]{8EA9DB} & \cellcolor[HTML]{8EA9DB} &  & \cellcolor[HTML]{8EA9DB} &  & \cellcolor[HTML]{BFBFBF} &  &  &  &  &  &  &  &  &  &  &  &  &  &  \\
\rowcolor[HTML]{F2F2F2} 
\cellcolor[HTML]{FCE4D6}S 7SVE V &  & \cellcolor[HTML]{8EA9DB} & \cellcolor[HTML]{F4B084} & \cellcolor[HTML]{F4B084} &  & \cellcolor[HTML]{F4B084} & \cellcolor[HTML]{F4B084} & \cellcolor[HTML]{F4B084} &  &  & \cellcolor[HTML]{F4B084} & \cellcolor[HTML]{F4B084} &  &  &  &  &  &  & \cellcolor[HTML]{BFBFBF} &  &  &  &  &  &  &  &  &  &  &  &  &  &  &  \\
\rowcolor[HTML]{F2F2F2} 
\cellcolor[HTML]{FCE4D6}E 7SVE V &  &  &  & \cellcolor[HTML]{F4B084} &  & \cellcolor[HTML]{F4B084} & \cellcolor[HTML]{F4B084} & \cellcolor[HTML]{F4B084} &  &  &  & \cellcolor[HTML]{F4B084} &  &  &  &  & \cellcolor[HTML]{F4B084} & \cellcolor[HTML]{BFBFBF} &  &  &  &  &  &  &  &  &  &  &  &  &  &  &  &  \\
\rowcolor[HTML]{F2F2F2} 
\cellcolor[HTML]{FCE4D6}C 7SVE V & \cellcolor[HTML]{8EA9DB} & \cellcolor[HTML]{8EA9DB} & \cellcolor[HTML]{8EA9DB} & \cellcolor[HTML]{8EA9DB} & \cellcolor[HTML]{8EA9DB} & \cellcolor[HTML]{8EA9DB} & \cellcolor[HTML]{8EA9DB} & \cellcolor[HTML]{8EA9DB} &  &  & \cellcolor[HTML]{8EA9DB} & \cellcolor[HTML]{8EA9DB} &  & \cellcolor[HTML]{8EA9DB} & \cellcolor[HTML]{8EA9DB} & \cellcolor[HTML]{8EA9DB} & \cellcolor[HTML]{BFBFBF} &  &  &  &  &  &  &  &  &  &  &  &  &  &  &  &  &  \\
\rowcolor[HTML]{F2F2F2} 
\cellcolor[HTML]{FCE4D6}SE 6SVE V &  &  &  &  &  & \cellcolor[HTML]{F4B084} & \cellcolor[HTML]{F4B084} & \cellcolor[HTML]{F4B084} &  &  &  &  &  &  &  & \cellcolor[HTML]{BFBFBF} &  &  &  &  &  &  &  &  &  &  &  &  &  &  &  &  &  &  \\
\rowcolor[HTML]{F2F2F2} 
\cellcolor[HTML]{FCE4D6}S 6SVE V &  &  &  &  &  & \cellcolor[HTML]{F4B084} & \cellcolor[HTML]{F4B084} &  &  &  &  &  &  &  & \cellcolor[HTML]{BFBFBF} &  &  &  &  &  &  &  &  &  &  &  &  &  &  &  &  &  &  &  \\
\rowcolor[HTML]{F2F2F2} 
\cellcolor[HTML]{FCE4D6}E 6SVE V &  &  &  &  &  & \cellcolor[HTML]{F4B084} &  &  &  &  &  &  &  & \cellcolor[HTML]{BFBFBF} &  &  &  &  &  &  &  &  &  &  &  &  &  &  &  &  &  &  &  &  \\
\rowcolor[HTML]{F2F2F2} 
\cellcolor[HTML]{FCE4D6}C 6SVE V &  & \cellcolor[HTML]{8EA9DB} & \cellcolor[HTML]{F4B084} & \cellcolor[HTML]{F4B084} &  & \cellcolor[HTML]{F4B084} & \cellcolor[HTML]{F4B084} & \cellcolor[HTML]{F4B084} &  &  & \cellcolor[HTML]{F4B084} & \cellcolor[HTML]{F4B084} & \cellcolor[HTML]{BFBFBF} &  &  &  &  &  &  &  &  &  &  &  &  &  &  &  &  &  &  &  &  &  \\
\rowcolor[HTML]{F2F2F2} 
\cellcolor[HTML]{FCE4D6}SE 5SVE V & \cellcolor[HTML]{F4B084} &  &  &  & \cellcolor[HTML]{8EA9DB} & \cellcolor[HTML]{F4B084} &  &  & \cellcolor[HTML]{8EA9DB} & \cellcolor[HTML]{8EA9DB} &  & \cellcolor[HTML]{BFBFBF} &  &  &  &  &  &  &  &  &  &  &  &  &  &  &  &  &  &  &  &  &  &  \\
\rowcolor[HTML]{F2F2F2} 
\cellcolor[HTML]{FCE4D6}S 5SVE V & \cellcolor[HTML]{F4B084} &  &  &  &  & \cellcolor[HTML]{F4B084} & \cellcolor[HTML]{F4B084} &  &  & \cellcolor[HTML]{F4B084} & \cellcolor[HTML]{BFBFBF} &  &  &  &  &  &  &  &  &  &  &  &  &  &  &  &  &  &  &  &  &  &  &  \\
\rowcolor[HTML]{F2F2F2} 
\cellcolor[HTML]{FCE4D6}E 5SVE V &  & \cellcolor[HTML]{8EA9DB} & \cellcolor[HTML]{F4B084} & \cellcolor[HTML]{F4B084} &  & \cellcolor[HTML]{F4B084} & \cellcolor[HTML]{F4B084} & \cellcolor[HTML]{F4B084} &  & \cellcolor[HTML]{BFBFBF} &  &  &  &  &  &  &  &  &  &  &  &  &  &  &  &  &  &  &  &  &  &  &  &  \\
\rowcolor[HTML]{F2F2F2} 
\cellcolor[HTML]{FCE4D6}C 5SVE V &  & \cellcolor[HTML]{F4B084} & \cellcolor[HTML]{F4B084} & \cellcolor[HTML]{F4B084} &  & \cellcolor[HTML]{F4B084} & \cellcolor[HTML]{F4B084} & \cellcolor[HTML]{F4B084} & \cellcolor[HTML]{BFBFBF} &  &  &  &  &  &  &  &  &  &  &  &  &  &  &  &  &  &  &  &  &  &  &  &  &  \\
\rowcolor[HTML]{F2F2F2} 
\cellcolor[HTML]{FCE4D6}SE 4SVE V & \cellcolor[HTML]{8EA9DB} &  &  &  & \cellcolor[HTML]{8EA9DB} &  &  & \cellcolor[HTML]{BFBFBF} &  &  &  &  &  &  &  &  &  &  &  &  &  &  &  &  &  &  &  &  &  &  &  &  &  &  \\
\rowcolor[HTML]{F2F2F2} 
\cellcolor[HTML]{FCE4D6}S 4SVE V & \cellcolor[HTML]{8EA9DB} &  &  &  & \cellcolor[HTML]{8EA9DB} &  & \cellcolor[HTML]{BFBFBF} &  &  &  &  &  &  &  &  &  &  &  &  &  &  &  &  &  &  &  &  &  &  &  &  &  &  &  \\
\rowcolor[HTML]{F2F2F2} 
\cellcolor[HTML]{FCE4D6}E 4SVE V & \cellcolor[HTML]{8EA9DB} &  &  &  & \cellcolor[HTML]{8EA9DB} & \cellcolor[HTML]{BFBFBF} &  &  &  &  &  &  &  &  &  &  &  &  &  &  &  &  &  &  &  &  &  &  &  &  &  &  &  &  \\
\rowcolor[HTML]{F2F2F2} 
\cellcolor[HTML]{FCE4D6}C 4SVE V &  & \cellcolor[HTML]{8EA9DB} & \cellcolor[HTML]{8EA9DB} & \cellcolor[HTML]{8EA9DB} & \cellcolor[HTML]{BFBFBF} &  &  &  &  &  &  &  &  &  &  &  &  &  &  &  &  &  &  &  &  &  &  &  &  &  &  &  &  &  \\
\rowcolor[HTML]{F2F2F2} 
\cellcolor[HTML]{FCE4D6}SE 3SVE V & \cellcolor[HTML]{F4B084} &  &  & \cellcolor[HTML]{BFBFBF} &  &  &  &  &  &  &  &  &  &  &  &  &  &  &  &  &  &  &  &  &  &  &  &  &  &  &  &  &  &  \\
\rowcolor[HTML]{F2F2F2} 
\cellcolor[HTML]{FCE4D6}S 3SVE V & \cellcolor[HTML]{F4B084} &  & \cellcolor[HTML]{BFBFBF} &  &  &  &  &  &  &  &  &  &  &  &  &  &  &  &  &  &  &  &  &  &  &  &  &  &  &  &  &  &  &  \\
\rowcolor[HTML]{F2F2F2} 
\cellcolor[HTML]{FCE4D6}E 3SVE V & \cellcolor[HTML]{F4B084} & \cellcolor[HTML]{BFBFBF} &  &  &  &  &  &  &  &  &  &  &  &  &  &  &  &  &  &  &  &  &  &  &  &  &  &  &  &  &  &  &  &  \\ \hline
\end{tabular}%
    }
\end{table}

\begin{table}[ht]
\caption{Shock Mann-Whitney Row/Col comparison}
\label{tab:shock_MW}
\resizebox{\columnwidth}{!}{%
\begin{tabular}{|c|ccccccccccccccccccccccccccccccc|}
\hline
\rowcolor[HTML]{D9E1F2} 
\cellcolor[HTML]{FFFFFF}\rotatebox[origin=c]{90}{} & \rotatebox[origin=c]{90}{E 3SVE V} & \rotatebox[origin=c]{90}{SE 3SVE V} & \rotatebox[origin=c]{90}{C 3SVE V} & \rotatebox[origin=c]{90}{S 3SVE V} & \rotatebox[origin=c]{90}{E 4SVE V} & \rotatebox[origin=c]{90}{SE 4SVE V} & \rotatebox[origin=c]{90}{S 4SVE V} & \rotatebox[origin=c]{90}{C 4SVE V} & \rotatebox[origin=c]{90}{E 5SVE V} & \rotatebox[origin=c]{90}{SE 5SVE V} & \rotatebox[origin=c]{90}{C 5SVE V} & \rotatebox[origin=c]{90}{S 5SVE V} & \rotatebox[origin=c]{90}{E 6SVE V} & \rotatebox[origin=c]{90}{SE 6SVE V} & \rotatebox[origin=c]{90}{C 6SVE V} & \rotatebox[origin=c]{90}{S 6SVE V} & \rotatebox[origin=c]{90}{E 7SVE V} & \rotatebox[origin=c]{90}{SE 7SVE V} & \rotatebox[origin=c]{90}{C 7SVE V} & \rotatebox[origin=c]{90}{S 7SVE V} & \rotatebox[origin=c]{90}{S 3 G} & {\color[HTML]{000000} \rotatebox[origin=c]{90}{S 6 G}} & \rotatebox[origin=c]{90}{S 8 G} & \rotatebox[origin=c]{90}{S 3 SP} & \rotatebox[origin=c]{90}{C 8 SP} & \rotatebox[origin=c]{90}{C 3 HP} & \rotatebox[origin=c]{90}{S 7 HP} & \rotatebox[origin=c]{90}{Rubber} & \rotatebox[origin=c]{90}{Coffee} & \rotatebox[origin=c]{90}{SEBS30} & \rotatebox[origin=c]{90}{S 6 G} \\ \hline
\rowcolor[HTML]{F2F2F2} 
\cellcolor[HTML]{FCE4D6}SEBS0 &  & \cellcolor[HTML]{F4B084} & \cellcolor[HTML]{8EA9DB} &  & \cellcolor[HTML]{8EA9DB} & \cellcolor[HTML]{8EA9DB} & \cellcolor[HTML]{8EA9DB} & \cellcolor[HTML]{8EA9DB} & \cellcolor[HTML]{8EA9DB} & \cellcolor[HTML]{8EA9DB} &  &  & \cellcolor[HTML]{8EA9DB} &  & \cellcolor[HTML]{8EA9DB} &  &  & \cellcolor[HTML]{8EA9DB} &  &  & \cellcolor[HTML]{F4B084} &  & \cellcolor[HTML]{F4B084} &  &  &  & \cellcolor[HTML]{F4B084} &  &  &  & \cellcolor[HTML]{F4B084} \\
\rowcolor[HTML]{8EA9DB} 
\cellcolor[HTML]{FCE4D6}SEBS30 &  & \cellcolor[HTML]{F4B084} &  &  &  &  &  &  &  &  &  &  &  &  &  &  &  &  &  &  &  & \cellcolor[HTML]{F2F2F2} & \cellcolor[HTML]{F2F2F2} &  &  &  & \cellcolor[HTML]{F4B084} &  &  & \cellcolor[HTML]{D9D9D9} & \cellcolor[HTML]{D9D9D9} \\
\cellcolor[HTML]{FCE4D6}Coffee & \cellcolor[HTML]{8EA9DB} & \cellcolor[HTML]{F4B084} & \cellcolor[HTML]{8EA9DB} & \cellcolor[HTML]{F2F2F2} & \cellcolor[HTML]{8EA9DB} & \cellcolor[HTML]{8EA9DB} & \cellcolor[HTML]{8EA9DB} & \cellcolor[HTML]{F2F2F2} & \cellcolor[HTML]{8EA9DB} & \cellcolor[HTML]{8EA9DB} & \cellcolor[HTML]{F2F2F2} & \cellcolor[HTML]{F2F2F2} & \cellcolor[HTML]{8EA9DB} & \cellcolor[HTML]{F2F2F2} & \cellcolor[HTML]{8EA9DB} & \cellcolor[HTML]{F2F2F2} & \cellcolor[HTML]{F2F2F2} & \cellcolor[HTML]{F4B084} & \cellcolor[HTML]{F2F2F2} & \cellcolor[HTML]{F2F2F2} & \cellcolor[HTML]{F4B084} & \cellcolor[HTML]{F4B084} & \cellcolor[HTML]{F2F2F2} & \cellcolor[HTML]{F2F2F2} & \cellcolor[HTML]{F4B084} & \cellcolor[HTML]{F4B084} & \cellcolor[HTML]{F4B084} & \cellcolor[HTML]{F4B084} & \cellcolor[HTML]{D9D9D9} & \cellcolor[HTML]{F2F2F2} & \cellcolor[HTML]{F2F2F2} \\
\rowcolor[HTML]{8EA9DB} 
\cellcolor[HTML]{FCE4D6}Rubber &  & \cellcolor[HTML]{F4B084} &  &  &  &  &  &  &  &  &  &  &  &  &  &  &  &  &  &  &  & \cellcolor[HTML]{F4B084} & \cellcolor[HTML]{F4B084} &  & \cellcolor[HTML]{F2F2F2} & \cellcolor[HTML]{F2F2F2} & \cellcolor[HTML]{F4B084} & \cellcolor[HTML]{D9D9D9} & \cellcolor[HTML]{F2F2F2} & \cellcolor[HTML]{F2F2F2} & \cellcolor[HTML]{F2F2F2} \\
\rowcolor[HTML]{8EA9DB} 
\cellcolor[HTML]{FCE4D6}S 7 HP &  & \cellcolor[HTML]{F2F2F2} &  &  &  &  &  &  &  &  &  &  &  &  &  &  &  &  &  &  &  &  &  &  &  &  & \cellcolor[HTML]{D9D9D9} & \cellcolor[HTML]{F2F2F2} & \cellcolor[HTML]{F2F2F2} & \cellcolor[HTML]{F2F2F2} & \cellcolor[HTML]{F2F2F2} \\
\rowcolor[HTML]{8EA9DB} 
\cellcolor[HTML]{FCE4D6}C 3 HP &  & \cellcolor[HTML]{F4B084} &  &  &  &  &  &  &  &  &  &  &  &  &  &  &  &  &  &  &  &  & \cellcolor[HTML]{F4B084} &  & \cellcolor[HTML]{F2F2F2} & \cellcolor[HTML]{D9D9D9} & \cellcolor[HTML]{F2F2F2} & \cellcolor[HTML]{F2F2F2} & \cellcolor[HTML]{F2F2F2} & \cellcolor[HTML]{F2F2F2} & \cellcolor[HTML]{F2F2F2} \\
\rowcolor[HTML]{8EA9DB} 
\cellcolor[HTML]{FCE4D6}C 8 SP &  & \cellcolor[HTML]{F4B084} &  &  &  &  &  &  &  &  &  &  &  &  &  &  &  &  &  &  &  & \cellcolor[HTML]{F4B084} & \cellcolor[HTML]{F4B084} &  & \cellcolor[HTML]{D9D9D9} & \cellcolor[HTML]{F2F2F2} & \cellcolor[HTML]{F2F2F2} & \cellcolor[HTML]{F2F2F2} & \cellcolor[HTML]{F2F2F2} & \cellcolor[HTML]{F2F2F2} & \cellcolor[HTML]{F2F2F2} \\
\rowcolor[HTML]{F2F2F2} 
\cellcolor[HTML]{FCE4D6}S 3 SP &  & \cellcolor[HTML]{F4B084} & \cellcolor[HTML]{F4B084} &  & \cellcolor[HTML]{F4B084} &  &  &  &  & \cellcolor[HTML]{F4B084} &  &  &  &  &  &  &  & \cellcolor[HTML]{F4B084} &  &  & \cellcolor[HTML]{F4B084} & \cellcolor[HTML]{F4B084} & \cellcolor[HTML]{F4B084} & \cellcolor[HTML]{D9D9D9} &  &  &  &  &  &  &  \\
\rowcolor[HTML]{8EA9DB} 
\cellcolor[HTML]{FCE4D6}S 8 G &  & \cellcolor[HTML]{F4B084} &  &  &  &  &  &  &  &  &  &  &  &  &  &  &  &  & \cellcolor[HTML]{F2F2F2} &  &  &  & \cellcolor[HTML]{D9D9D9} & \cellcolor[HTML]{F2F2F2} & \cellcolor[HTML]{F2F2F2} & \cellcolor[HTML]{F2F2F2} & \cellcolor[HTML]{F2F2F2} & \cellcolor[HTML]{F2F2F2} & \cellcolor[HTML]{F2F2F2} & \cellcolor[HTML]{F2F2F2} & \cellcolor[HTML]{F2F2F2} \\
\rowcolor[HTML]{8EA9DB} 
\cellcolor[HTML]{FCE4D6}S 6 G &  & \cellcolor[HTML]{F4B084} &  &  &  &  &  &  &  &  &  &  &  &  &  &  &  &  &  &  &  & \cellcolor[HTML]{D9D9D9} & \cellcolor[HTML]{F2F2F2} & \cellcolor[HTML]{F2F2F2} & \cellcolor[HTML]{F2F2F2} & \cellcolor[HTML]{F2F2F2} & \cellcolor[HTML]{F2F2F2} & \cellcolor[HTML]{F2F2F2} & \cellcolor[HTML]{F2F2F2} & \cellcolor[HTML]{F2F2F2} & \cellcolor[HTML]{F2F2F2} \\
\rowcolor[HTML]{F2F2F2} 
\cellcolor[HTML]{FCE4D6}S 3 G & \cellcolor[HTML]{8EA9DB} &  &  & \cellcolor[HTML]{8EA9DB} &  &  &  & \cellcolor[HTML]{8EA9DB} & \cellcolor[HTML]{8EA9DB} &  &  & \cellcolor[HTML]{8EA9DB} & \cellcolor[HTML]{8EA9DB} & \cellcolor[HTML]{8EA9DB} & \cellcolor[HTML]{8EA9DB} & \cellcolor[HTML]{8EA9DB} & \cellcolor[HTML]{8EA9DB} &  & \cellcolor[HTML]{8EA9DB} & \cellcolor[HTML]{8EA9DB} & \cellcolor[HTML]{D9D9D9} &  &  &  &  &  &  &  &  &  &  \\
\rowcolor[HTML]{F2F2F2} 
\cellcolor[HTML]{FCE4D6}S 7SVE V &  & \cellcolor[HTML]{F4B084} & \cellcolor[HTML]{F4B084} &  &  &  &  &  &  & \cellcolor[HTML]{F4B084} &  &  &  &  &  &  &  & \cellcolor[HTML]{F4B084} &  & \cellcolor[HTML]{D9D9D9} &  &  &  &  &  &  &  &  &  &  &  \\
\rowcolor[HTML]{F2F2F2} 
\cellcolor[HTML]{FCE4D6}C 7SVE V & \cellcolor[HTML]{8EA9DB} & \cellcolor[HTML]{F4B084} & \cellcolor[HTML]{F4B084} &  & \cellcolor[HTML]{8EA9DB} &  & \cellcolor[HTML]{8EA9DB} &  &  & \cellcolor[HTML]{8EA9DB} &  &  & \cellcolor[HTML]{8EA9DB} &  &  &  &  & \cellcolor[HTML]{F4B084} & \cellcolor[HTML]{D9D9D9} &  &  &  &  &  &  &  &  &  &  &  &  \\
\rowcolor[HTML]{F2F2F2} 
\cellcolor[HTML]{FCE4D6}SE 7SVE V & \cellcolor[HTML]{8EA9DB} &  &  & \cellcolor[HTML]{8EA9DB} &  &  &  & \cellcolor[HTML]{8EA9DB} & \cellcolor[HTML]{8EA9DB} &  & \cellcolor[HTML]{8EA9DB} & \cellcolor[HTML]{8EA9DB} & \cellcolor[HTML]{8EA9DB} & \cellcolor[HTML]{8EA9DB} & \cellcolor[HTML]{8EA9DB} & \cellcolor[HTML]{8EA9DB} & \cellcolor[HTML]{8EA9DB} & \cellcolor[HTML]{D9D9D9} &  &  &  &  &  &  &  &  &  &  &  &  &  \\
\rowcolor[HTML]{F2F2F2} 
\cellcolor[HTML]{FCE4D6}E 7SVE V &  & \cellcolor[HTML]{F4B084} & \cellcolor[HTML]{F4B084} &  & \cellcolor[HTML]{F4B084} &  &  &  &  & \cellcolor[HTML]{F4B084} &  &  &  &  &  &  & \cellcolor[HTML]{D9D9D9} &  &  &  &  &  &  &  &  &  &  &  &  &  &  \\
\rowcolor[HTML]{F2F2F2} 
\cellcolor[HTML]{FCE4D6}S 6SVE V &  & \cellcolor[HTML]{F4B084} &  &  &  &  &  &  &  &  &  &  &  &  &  & \cellcolor[HTML]{D9D9D9} &  &  &  &  &  &  &  &  &  &  &  &  &  &  &  \\
\rowcolor[HTML]{F2F2F2} 
\cellcolor[HTML]{FCE4D6}C 6SVE V &  & \cellcolor[HTML]{F4B084} &  &  &  &  &  &  &  &  &  &  &  &  & \cellcolor[HTML]{D9D9D9} &  &  &  &  &  &  &  &  &  &  &  &  &  &  &  &  \\
\rowcolor[HTML]{F2F2F2} 
\cellcolor[HTML]{FCE4D6}SE 6SVE V &  & \cellcolor[HTML]{F4B084} & \cellcolor[HTML]{F4B084} &  & \cellcolor[HTML]{F4B084} &  &  &  &  & \cellcolor[HTML]{F4B084} &  &  &  & \cellcolor[HTML]{D9D9D9} &  &  &  &  &  &  &  &  &  &  &  &  &  &  &  &  &  \\
\rowcolor[HTML]{F2F2F2} 
\cellcolor[HTML]{FCE4D6}E 6SVE V &  & \cellcolor[HTML]{F4B084} &  &  &  &  &  &  &  &  &  &  & \cellcolor[HTML]{D9D9D9} &  &  &  &  &  &  &  &  &  &  &  &  &  &  &  &  &  &  \\
\rowcolor[HTML]{F2F2F2} 
\cellcolor[HTML]{FCE4D6}S 5SVE V &  & \cellcolor[HTML]{F4B084} & \cellcolor[HTML]{F4B084} &  &  &  &  &  &  &  &  & \cellcolor[HTML]{D9D9D9} &  &  &  &  &  &  &  &  &  &  &  &  &  &  &  &  &  &  &  \\
\rowcolor[HTML]{F2F2F2} 
\cellcolor[HTML]{FCE4D6}C 5SVE V &  & \cellcolor[HTML]{F4B084} &  &  &  &  &  &  &  &  & \cellcolor[HTML]{D9D9D9} &  &  &  &  &  &  &  &  &  &  &  &  &  &  &  &  &  &  &  &  \\
\rowcolor[HTML]{F2F2F2} 
\cellcolor[HTML]{FCE4D6}SE 5SVE V &  &  &  & \cellcolor[HTML]{8EA9DB} &  &  &  &  &  & \cellcolor[HTML]{D9D9D9} &  &  &  &  &  &  &  &  &  &  &  &  &  &  &  &  &  &  &  &  &  \\
\rowcolor[HTML]{F2F2F2} 
\cellcolor[HTML]{FCE4D6}E 5SVE V &  & \cellcolor[HTML]{F4B084} & \cellcolor[HTML]{F4B084} &  &  &  &  &  & \cellcolor[HTML]{D9D9D9} &  &  &  &  &  &  &  &  &  &  &  &  &  &  &  &  &  &  &  &  &  &  \\
\rowcolor[HTML]{F2F2F2} 
\cellcolor[HTML]{FCE4D6}C 4SVE V &  & \cellcolor[HTML]{F4B084} &  &  &  &  &  & \cellcolor[HTML]{D9D9D9} &  &  &  &  &  &  &  &  &  &  &  &  &  &  &  &  &  &  &  &  &  &  &  \\
\rowcolor[HTML]{F2F2F2} 
\cellcolor[HTML]{FCE4D6}S 4SVE V &  & \cellcolor[HTML]{F4B084} &  &  &  &  & \cellcolor[HTML]{D9D9D9} &  &  &  &  &  &  &  &  &  &  &  &  &  &  &  &  &  &  &  &  &  &  &  &  \\
\rowcolor[HTML]{F2F2F2} 
\cellcolor[HTML]{FCE4D6}SE 4SVE V &  & \cellcolor[HTML]{F4B084} &  &  &  & \cellcolor[HTML]{D9D9D9} &  &  &  &  &  &  &  &  &  &  &  &  &  &  &  &  &  &  &  &  &  &  &  &  &  \\
\rowcolor[HTML]{F2F2F2} 
\cellcolor[HTML]{FCE4D6}E 4SVE V &  & \cellcolor[HTML]{F4B084} &  &  & \cellcolor[HTML]{D9D9D9} &  &  &  &  &  &  &  &  &  &  &  &  &  &  &  &  &  &  &  &  &  &  &  &  &  &  \\
\rowcolor[HTML]{F2F2F2} 
\cellcolor[HTML]{FCE4D6}S 3SVE V &  & \cellcolor[HTML]{F4B084} &  & \cellcolor[HTML]{D9D9D9} &  &  &  &  &  &  &  &  &  &  &  &  &  &  &  &  &  &  &  &  &  &  &  &  &  &  &  \\
\rowcolor[HTML]{F2F2F2} 
\cellcolor[HTML]{FCE4D6}C 3SVE V & \cellcolor[HTML]{8EA9DB} & \cellcolor[HTML]{F4B084} & \cellcolor[HTML]{D9D9D9} &  &  &  &  &  &  &  &  &  &  &  &  &  &  &  &  &  &  &  &  &  &  &  &  &  &  &  &  \\
\rowcolor[HTML]{F2F2F2} 
\cellcolor[HTML]{FCE4D6}SE 3SVE V & \cellcolor[HTML]{8EA9DB} & \cellcolor[HTML]{D9D9D9} &  &  &  &  &  &  &  &  &  &  &  &  &  &  &  &  &  &  &  &  &  &  &  &  &  &  &  &  &  \\
\rowcolor[HTML]{F2F2F2} 
\cellcolor[HTML]{FCE4D6}{SE 3SVE V} &  & \cellcolor[HTML]{D9D9D9} &  &  &  &  &  &  &  &  &  &  &  &  &  &  &  &  &  &  &  &  &  &  &  &  &  &  &  &  &  \\ \hline
\end{tabular}%
}
\end{table}

\end{document}